\documentclass{article} 
\usepackage{times}
\usepackage{microtype}
\usepackage{booktabs} 
\usepackage{ragged2e}
\usepackage{soul}

\usepackage{graphicx}
\usepackage{caption}
\usepackage{subcaption}

\usepackage[accepted]{Styles/icml2025}

\usepackage{amsmath,amsfonts,bm}

\def\eqref#1{equation~\ref{#1}}

\def\1{\bm{1}}

\DeclareMathAlphabet{\mathsfit}{\encodingdefault}{\sfdefault}{m}{sl}
\SetMathAlphabet{\mathsfit}{bold}{\encodingdefault}{\sfdefault}{bx}{n}

\usepackage{hyperref}
\usepackage{url}
\usepackage{booktabs}       
\usepackage{amsfonts}       
\usepackage{nicefrac}       
\usepackage{microtype}      
\usepackage{xcolor}         

\sethlcolor{yellow}

\begin{document}

\twocolumn[
\icmltitle{Which Attention Heads Matter for  In-Context Learning?}



\icmlsetsymbol{equal}{*}

\begin{icmlauthorlist}
\icmlauthor{Kayo Yin}{yyy}
\icmlauthor{Jacob Steinhardt}{yyy}
\end{icmlauthorlist}

\icmlaffiliation{yyy}{UC Berkeley}

\icmlcorrespondingauthor{Kayo Yin}{kayoyin@berkeley.edu}
\icmlcorrespondingauthor{Jacob Steinhardt}{jsteinhardt@berkeley.edu}

\icmlkeywords{interpretability, in-context learning}

\vskip 0.3in
]



\printAffiliationsAndNotice{}  

\begin{abstract}
Large language models (LLMs) exhibit impressive in-context learning (ICL) capability, enabling them to perform new tasks using only a few demonstrations in the prompt. 
Two different mechanisms have been proposed to explain ICL:
induction heads that find and copy relevant tokens, and function vector (FV) heads whose activations compute a latent encoding of the ICL task.
To better understand which of the two distinct mechanisms drives ICL, we study and compare induction heads and FV heads in 12 language models. 

Through detailed ablations, we discover that few-shot ICL performance depends primarily on FV heads, especially in larger models. In addition, we uncover that FV and induction heads are connected: many FV heads
start as induction heads during training before transitioning to the FV mechanism. This leads us to speculate that induction facilitates learning the more complex FV mechanism that ultimately drives ICL\footnotemark[2]. 
\end{abstract}
\footnotetext[2]{Code and data: \url{https://github.com/kayoyin/icl-heads}.}

\section{Introduction}

One of the most remarkable features of large language models (LLM) is their ability to perform in-context learning (ICL), where they can adapt to various new tasks using only a few demonstrations at inference time. This capability has become a crucial tool for adapting pre-trained LLMs to specific tasks, sparking significant research interest in understanding its underlying mechanisms \citep{ih, akyurek2022learning, von2023transformers}.

To date, two key mechanisms have been primarily associated with ICL, substantiated by different lines of evidence. First, \textit{induction circuits} \citep{elhage2021mathematical} were hypothesized to be the primary mechanism behind ICL in LLMs \citep{ih, singh2024needs, crosbie2024induction, dong2022survey}. Induction circuits operate by identifying previous occurrences of the current token in the prompt and copying the subsequent token. More recently, \citet{fv} and \citet{tv} propose the existence of \textit{function vectors} (FV). FVs are a compact representation of a task extracted from specific attention heads, and they can be added to a model's computation to recover ICL behavior without in-context demonstrations.

\begin{figure}[tbp]
\captionsetup[subfigure]{justification=Centering}
\begin{subfigure}{\linewidth}
    \includegraphics[width=0.9\linewidth]{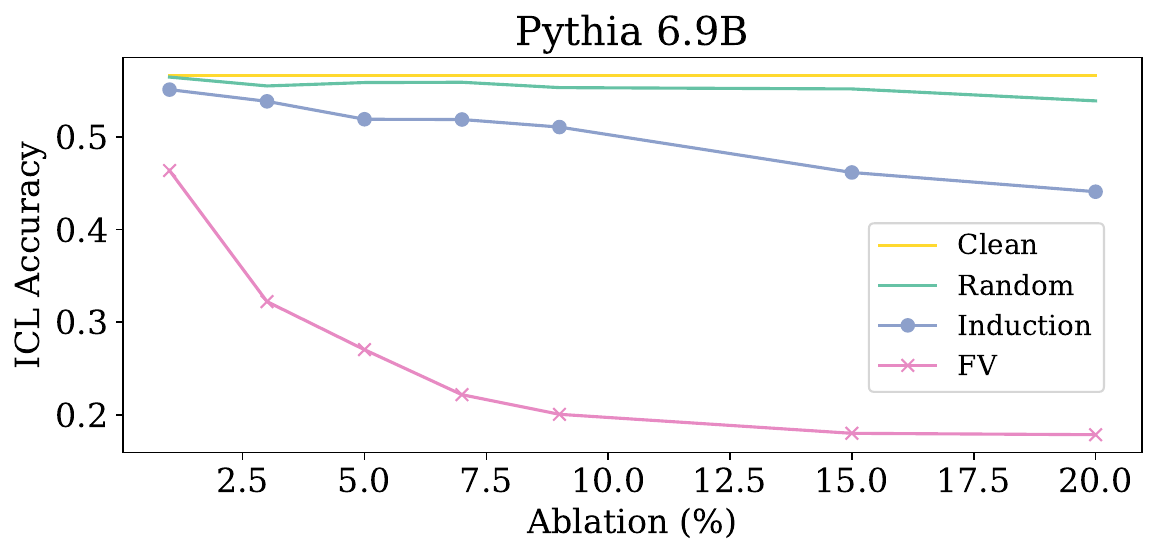}
    \caption{ICL accuracy of Pythia 6.9B across different percentages of heads ablated.}
    \label{fig:abl_6}
\end{subfigure} 
\begin{subfigure}{\linewidth}
    \includegraphics[width=\linewidth]{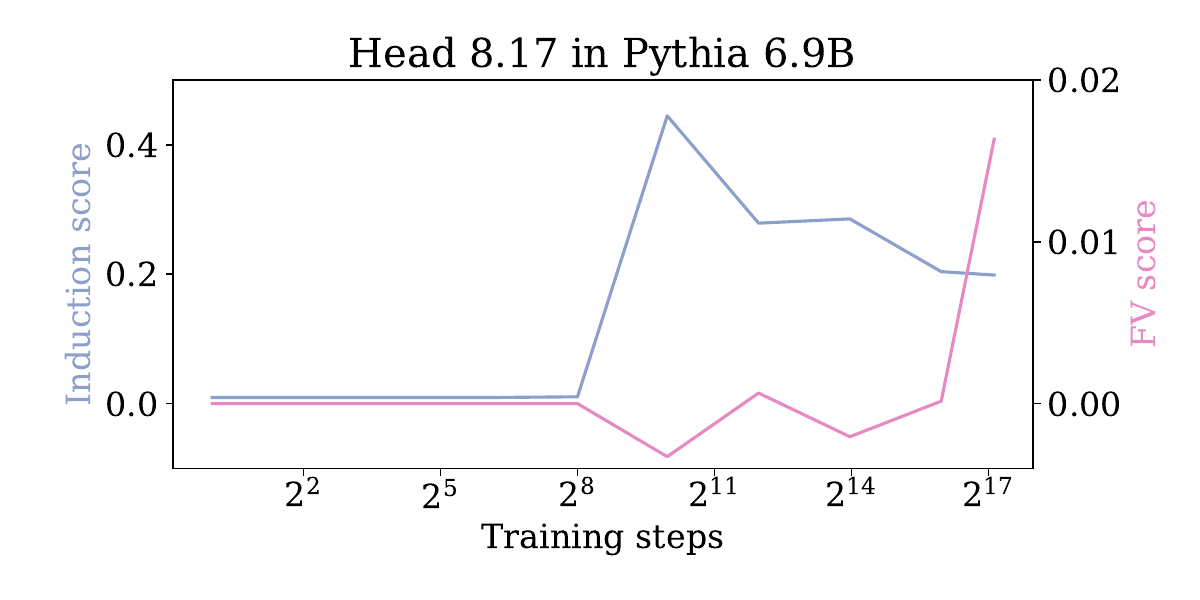}
    \caption{Induction score (blue) and FV score (pink) of an FV head during training.}
    \label{fig:indv_ckpt_6}
\end{subfigure}
\caption{(a) Ablating function vector (FV) heads significantly degrades few-shot in-context learning (ICL) accuracy, while ablating induction heads has minimal impact beyond ablating random heads. (b) Evolution of an FV head during training, demonstrating high induction scores earlier in training that decrease as FV score emerges. This pattern suggests induction may serve as a precursor for FV mechanism.}
\label{fig:fig1}
\end{figure}

\begin{table*}[htp]
    \centering
    \caption{Summary of findings in this work, where $\checkmark$ represents findings with evidence directly shown by our experiments and $\sim$ represents conjectures that our results suggest.}
      \resizebox{\linewidth}{!}{
    \begin{tabular}{lccl}
    \toprule
     \textbf{Findings} & \textbf{Evidence}  & \textbf{Section} & \textbf{Contribution}\\
     \midrule
Induction heads and FV heads are distinct. & $\checkmark$ & \ref{sec:overlap} & Context-setting\\
Induction scores and FV scores are correlated. & $\checkmark$ & \ref{sec:overlap} & Context-setting\\
Ablating FV heads hurts few-shot ICL accuracy more than ablating induction heads. & $\checkmark$ & \ref{sec:abl} & Main finding \\
Some FV heads evolve from induction heads during training. & $\checkmark$ & \ref{sec:ckpt} & Main finding \\
FV heads implement more complex or abstract computations than induction heads. & $\sim$ &  \ref{sec:ckpt} & Speculation \\
\bottomrule
    \end{tabular}}
    \label{tab:hypotheses}
\end{table*}


To resolve whether one or both mechanisms drive ICL in transformer LLMs, we conduct a comprehensive study of the attention heads implementing these mechanisms (termed \textit{induction heads} and \textit{FV heads}) across 12 decoder-only transformer models ranging from 70M to 7B parameters (Table \ref{tab:models}) and 45 natural language ICL tasks (listed in Appendix \ref{app:tasks}). Our analysis reveals several key findings.

First, we verify that there is a difference to explain, i.e. that induction and FV heads are indeed distinct (\S\ref{sec:overlap}). Across models, there is a low or zero overlap between induction and FV heads. These heads also have distinct characteristics: induction heads generally appear in slightly earlier layers than FV heads, and emerge significantly earlier during training. On the other hand, there are correlations in behavior: FV heads behave more similarly to induction heads than a random head from the same network, and vice versa.

Second, through ablation studies (\S\ref{sec:abl}), we demonstrate that FV heads are the primary drivers of ICL performance. Removing FV heads substantially degrades ICL task accuracy, while removing induction heads has a limited effect (Figure \ref{fig:abl_6}). This effect is consistent in all 12 models we studied, and becomes more pronounced in larger models (\S\ref{sec:abl}). Interestingly, this challenges the prevailing view of induction heads as the key mechanism for few-shot ICL \citep{ih, crosbie2024induction, dong2022survey}.

Third, we reconcile our findings with previous work by identifying three key methodological differences (\S\ref{sec:bg}): earlier studies used a different metric for ICL that does not strongly track few-shot performance, did not account for correlations between FV and induction heads, and sometimes focused on small models. We find the choice of metric is the most significant factor, and discuss this in detail in \S\ref{sec:bg} with detailed experiments in \S\ref{sec:abl}.

Finally, by analyzing training dynamics (\S\ref{sec:ckpt}), we uncover a surprising developmental relationship: many induction heads \textit{evolve} into FV heads during training, but the reverse never occurs (Figure \ref{fig:indv_ckpt_6}). 
This leads us to speculate that induction heads facilitate learning the more complex FV heads for ICL -- the FV mechanism is more effective at performing ICL, and therefore eventually replaces the simpler induction mechanism \S\ref{sec:disc}. 

Aside from clarifying the drivers of few-shot ICL, our findings offer broader lessons for model interpretability research. They highlight how correlations between related mechanisms can lead to illusory explanations (e.g. the confounding effect of the correlation between induction and FV heads), and the choice of definitions may lead to different conclusions (e.g. different metrics to measure ICL).
Additionally, our results challenge strong versions of universality -- the difference between the importance of FV heads and induction heads shifts significantly with model scale (\S\ref{sec:abl}). 

\section{Background \& related work}\label{sec:bg}

We present a comparative analysis of two mechanisms proposed to explain ICL: induction heads \citep{elhage2021mathematical, ih} and FV heads \citep{fv, tv}.

\subsection{Induction heads}

Induction heads were first identified by \citet{elhage2021mathematical} and extensively studied by \citet{ih} as the mechanism behind ICL. They are attention heads that operate by identifying repeated patterns in the input: when processing a token, they attend to the token that followed a previous occurrence of the same token, predicting it will appear next.

The initial evidence for induction heads' role in ICL came from \citet{ih}, who studied small attention-only models (1-3 layers). They observed that the emergence of induction heads during training coincided with improvements in ICL ability -- measured as the difference between the loss at the 500th versus 50th token in the context. Their ablation studies showed that removing induction heads impaired this metric.

To identify and analyze induction heads, we measure their \textbf{induction scores} using the TransformerLens framework \citep{tflens}. For each attention head $a$, we compute its induction score on a synthetic sequence constructed by repeating a uniformly sampled random token sequence: $r = r_1r_2...r_{50}r'_1r'_2...r'_{50}$. The induction score is defined as:
$$
S_I(a, r) = \sum_{i=1}^{50} a_{r'_i \rightarrow r_{i+1}}
$$
where $a_{r'_i \rightarrow r_{i+1}}$ represents the attention weight that head $a$ places on token $r_{i+1}$ when processing token $r'_i$. For each attention head in each model, we take the mean induction score over 1000 samples of random sequences $r$, normalized by total attention mass to obtain a value between 0 and 1. 

\subsection{FV heads}

Function vectors (FV) were concurrently discovered by \citet{fv} and \citet{tv}. They represent a different mechanism for ICL: FVs are compact vector representations of ICL tasks that can be extracted from specific attention heads and added back into the language model's computations to reproduce ICL behavior.
We refer to the attention heads that encode and transport these function vectors as \textbf{FV heads}.



To identify FV heads, we employ the casual mediation analysis framework from \citet{fv}. For each ICL task $t$ in our task set $\mathcal{T}$, where $t$ is defined by a dataset $P_t$ of in-context prompts $p_i^t \in P_t$ consisting of input-output pairs $(x_i,y_i)$, we:
\begin{enumerate}
    \item Compute the mean activation of an attention head $a$ over prompts in $P_t$:  $\bar{a}^t = \frac{1}{P_t}\sum_{p_i^t \in P_t}a(p_i^t)$
    \item Create corrupted ICL prompts $\tilde{p}_i^t \in \tilde{P}_t$ by randomly shuffling the output labels $\tilde{y_i}$ while maintaining the same inputs $x_i$
    \item Measure each head's \textbf{function vector score} (FV score) as its causal contribution to recovering the correct output $y$ for the input $x$ given corrupted examples $(x_i, \tilde{y_i})$ when its activation pattern is replaced with the mean task-conditioned activation $\bar{a}^t$:
    $$
S_{FV}(a | \tilde{p}_i^t) = f(\tilde{p}_i^t | a := \bar{a}^t)[y] - f(\tilde{p}_i^t)[y].
$$
\end{enumerate} 


For each attention head, we take the mean FV score across 37 natural language ICL tasks from \citep{fv} (Appendix \ref{app:tasks}), using 100 prompts per task. Each prompt contains 10 input-output demonstration pairs followed by a single test instance.


\subsection{Reconciling divergent findings}

While both mechanisms have been proposed by their respective works as the mechanism behind ICL, our side-by-side analysis of induction and FV heads reveals that FV heads seem to primarily contribute to ICL performance. 
We believe that the main reason for the divergence between our result and previous work lies in several intuitively related concepts in the literature that are assumed to be the same. 

First, there is an important distinction between two different conceptualizations of ICL:
\begin{itemize}
    \item On one hand, ICL is often used synonymously with few-shot learning from the prompt without parameter updates \citep{brown2020language, dong2022survey, wei2023larger}. We adopt this conceptualization of ICL in this paper since it is the most standard in the literature, and for clarity, we will use ``in-context learning'' in this paper with this definition. 
    \item On the other hand, ICL performance is measured in \citet{ih} by computing the difference between the model loss of the 500th token in the context and the loss of the 50th token. This difference was previously called ``ICL score'' but we recommend adopting distinct terminology to avoid confusion, and we call this \textbf{token-loss difference}. 
\end{itemize}

Our experiments reveal that these metrics capture different phenomena: FV heads strongly influence few-shot ICL accuracy but not token-loss difference, while induction heads show the opposite pattern (\S\ref{sec:abl}). This divergence of token-loss difference from few-shot ICL performance accounts for much of the apparent contradiction with previous findings.

Second, we find that induction heads and FV heads are correlated (\S\ref{sec:overlap}), a confound not previously controlled for. Initial ablation studies, including ours (\S\ref{sec:abl}), showed that removing induction heads significantly degrades few-shot ICL accuracy \citep{crosbie2024induction, bansal-etal-2023-rethinking}. However, when we control for this correlation by only ablating induction heads with low FV scores, their impact becomes comparable to random ablation. In contrast, ablating FV heads with low induction scores still significantly degrades ICL performance. This suggests that previous studies may have attributed to induction heads effects that actually stemmed from FV-like behavior in a subset of induction heads.

Third, scale matters. Previous work establishing induction heads as the key ICL mechanism \citep{ih, singh2024needs} focused on small models to enable detailed mechanistic analysis. However, we find that the relative importance of FV heads increases with model scale. In our smallest model (70M parameters), induction and FV heads have similar causal effects on few-shot ICL (\S\ref{app:abl}), but this does not hold true in larger models, which highlights the importance of studying these phenomena across different model scales.




\begin{table*}[htp]
    \centering
    \caption{Models studied in this work. We use huggingface implementations \citep{huggingface} for all models. We report the number of parameters, number of layers $|L|$, total number of attention heads $|a|$, and the dimension of each head dim$_a$ for each model.}
    \vspace{10pt}
      \resizebox{\linewidth}{!}{
    \begin{tabular}{llcccc}
    \toprule
     \textbf{Model} & \textbf{Huggingface ID}  & \textbf{Parameters} & $|L|$ & $|a|$ & dim$_a$\\
     \midrule
Pythia \citep{pythia}& \texttt{EleutherAI/pythia-70m-deduped}& 70M& 6&  48& 64\\
Pythia \citep{pythia}& \texttt{EleutherAI/pythia-160m-deduped}& 160M& 12&  144& 64\\
Pythia \citep{pythia}& \texttt{EleutherAI/pythia-410m-deduped}& 410M& 24&  384& 64\\
Pythia \citep{pythia}&\texttt{EleutherAI/pythia-1b-deduped}& 1B& 16&  128& 256\\   
Pythia \citep{pythia}& \texttt{EleutherAI/pythia-1.4b-deduped}& 1.4B& 24&  384& 128\\
Pythia \citep{pythia}& \texttt{EleutherAI/pythia-2.8b-deduped}& 2.8B& 32&  1024& 80\\
Pythia \citep{pythia}& \texttt{EleutherAI/pythia-6.9b-deduped}& 6.9B& 32&  1024& 128\\
\midrule
GPT-2 \citep{gpt2}& \texttt{openai-community/gpt2}& 117M& 12&  144& 64\\
GPT-2 \citep{gpt2}& \texttt{openai-community/gpt2-medium}& 345M& 24&  384& 64\\
GPT-2 \citep{gpt2}& \texttt{openai-community/gpt2-large}& 774M& 36&  720& 64\\
GPT-2 \citep{gpt2}& \texttt{openai-community/gpt2-xl}& 1.6B& 48&  1200& 64\\
\midrule
Llama 2 \citep{llama}& \texttt{meta-llama/Llama-2-7b-hf}& 7B& 32&  1024& 128\\
\bottomrule
    \end{tabular}}
    \label{tab:models}
\end{table*}

\section{Induction heads and function vector heads are distinct but correlated}\label{sec:overlap}

Before analyzing the relative contributions of induction and FV heads to ICL performance, we first establish that these represent distinct mechanisms, while noting important correlations between them.

\subsection{Head locations}
\begin{figure*}
\centering
    \includegraphics[width=\linewidth]{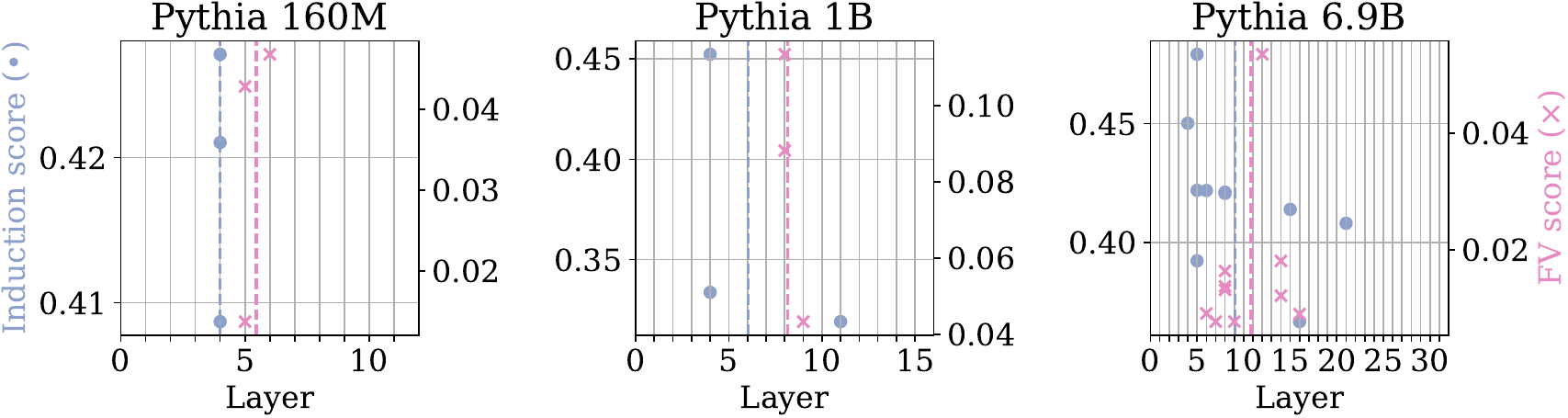}
  \caption{Location of induction heads (blue) and FV heads (pink) in model layers.
The average layer of induction and FV heads are shown in blue and pink dotted lines respectively. Most induction heads appear in early-middle layers, FV heads appear at layers slightly deeper than induction heads.}
  \label{fig:head_layers}
\end{figure*}

We begin by examining the location of the top induction and FV heads within the models. Figure \ref{fig:head_layers} shows the layers where the top 2\%\footnote{In certain cases, we need to differentiate between meaningful induction / FV heads and the long tail of attention heads that perform neither induction nor the FV mechanism. We choose the top 2\% induction and FV heads as the representative set of induction and FV heads, following \citet{fv}.} induction heads and FV heads appear in three representative Pythia models (see Appendix \ref{app:layers} for all 12 models).

In general, induction heads appear in early-middle layers and FV heads appear in slightly deeper layers than induction heads. This suggests that induction and FV heads do not fully overlap and are indeed distinct mechanisms. Moreover, the deeper locations of FV heads may indicate they implement more abstract computations than induction heads, though this interpretation remains speculative.

\subsection{Overlap between induction and FV heads}\label{sec:overlap2}

To further examine how distinct induction and FV heads are, we analyze the extent of the overlap between the two types of heads in two ways.

\begin{figure*}
\centering
    \includegraphics[width=0.95\linewidth]{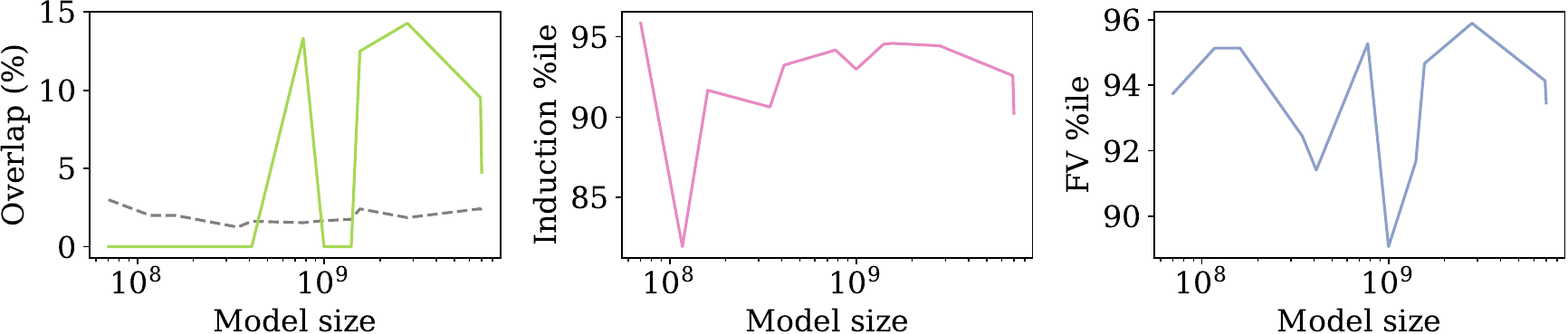}
  \caption{Percentage of head overlap between induction and FV heads (left) in green, and between induction and randomly sampled heads in gray. Percentile of induction score of FV heads (center). Percentile of FV score of induction heads (right). There is little overlap between induction and FV heads, but FV heads have relatively high induction scores and vice versa.}
  \label{fig:head_percentiles}
\end{figure*}

First, we measure direct overlap - the percentage of heads that rank in the top 2\% for both mechanisms: $100 \times \frac{|IH \cap FV|}{|IH|}$ where $IH$ and $FV$ represent the sets of top induction and FV heads respectively. The results show minimal overlap: seven of our twelve models show zero overlap, with the remaining models showing only 5-15\% overlap (Figure \ref{fig:head_percentiles} left). This leads us to conclude that \textbf{induction heads and FV heads are mostly distinct} and justifies studying them as two different mechanisms.

However, a more nuanced pattern emerges when we compute the percentile of the induction score of the top 2\% FV heads (Figure \ref{fig:head_percentiles} center) and the percentile of the FV score of the top 2\% induction heads (Figure \ref{fig:head_percentiles} right). In most models, FV heads are at around the 90-95th percentile of induction scores, and vice versa. Therefore, although there is little overlap between the sets of induction and FV heads, \textbf{induction and FV scores are correlated}: FV heads have high induction scores relative to other attention heads, and induction heads have relatively high FV scores
\footnote{In our main analysis, we do not rely on the correlation between the distribution of induction scores and FV scores across the full set of attention heads because there is a long tail of attention heads with low scores on both induction and FV. For completeness, we plot the induction and FV scores of all heads in Appendix \ref{app:corr}.}.

\section{Function vector heads drive in-context learning}\label{sec:abl}

Having established that induction and FV heads represent distinct mechanisms, we now investigate their relative causal importance for ICL through systematic ablation studies. Our analysis focuses primarily on few-shot ICL accuracy while also examining effects on token-loss difference for comparison with previous work.

\subsection{Method}


\textbf{Ablation.} To assess the causal contribution of different attention heads, we measure how ICL performance changes when specific heads are disabled. We use mean ablation, where we replace each target head's output with its average output across our task dataset (described in later sections). This approach avoids the out-of-distribution effects associated with zero ablation \citep{hase2021out,wang2023interpretability,zhang2024towards}, though our findings remain robust across different ablation methods (Appendix \ref{app:zero_ablate}).

To control for the correlation between induction and FV heads identified in Section \ref{sec:overlap}, we introduce ablation with ``exclusion'': when ablating $n$ FV heads, we select the top $n$ heads by FV score that are not in the top 2\% by induction score, and vice versa. This helps isolate the unique contributions of each mechanism.


\textbf{Few-shot ICL accuracy.}
Our primary metric evaluates ICL performance on a series of few-shot ICL tasks. Each ICL task is defined by a set of input-output pairs $(x_i, y_i)$. The model is prompted with 10 input-output exemplar pairs that demonstrate this task, and one query input $x_q$ that corresponds to a target output $y_q$ that is not part of the model's prompt. We compute the model's accuracy in predicting the correct output $y_q$. We summarize the full set of ICL tasks we study in Appendix \ref{app:tasks}.

To avoid leakage between ICL tasks used to identify FV heads and those used to evaluate FV head ablations, we randomly split the 37 ICL tasks from \citet{fv} into 26 tasks used to measure FV scores of heads, and 11 tasks to evaluate ICL performance. We also add 8 new tasks for ICL evaluation: 4 tasks are variations of tasks in \citet{fv}, and 4 are binding tasks from \citet{binding}. In total, we evaluate ICL accuracy on 19 natural language tasks, with 100 prompts per task.

\textbf{Token-loss difference.}
To compare with previous work, we also study the effect of ablations on token-loss difference used in \citet{ih}. We measure token-loss difference by taking the loss of the 50th token in the context minus the loss of the 500th token in the context\footnote{We invert the difference used in \citet{ih} so higher scores indicate better ICL performance.}, averaged over 10,000 examples from the Pile dataset \citep{pile}.  




\subsection{Results}

\begin{figure*}
\centering
\vspace{-30pt}
    \includegraphics[width=\linewidth]{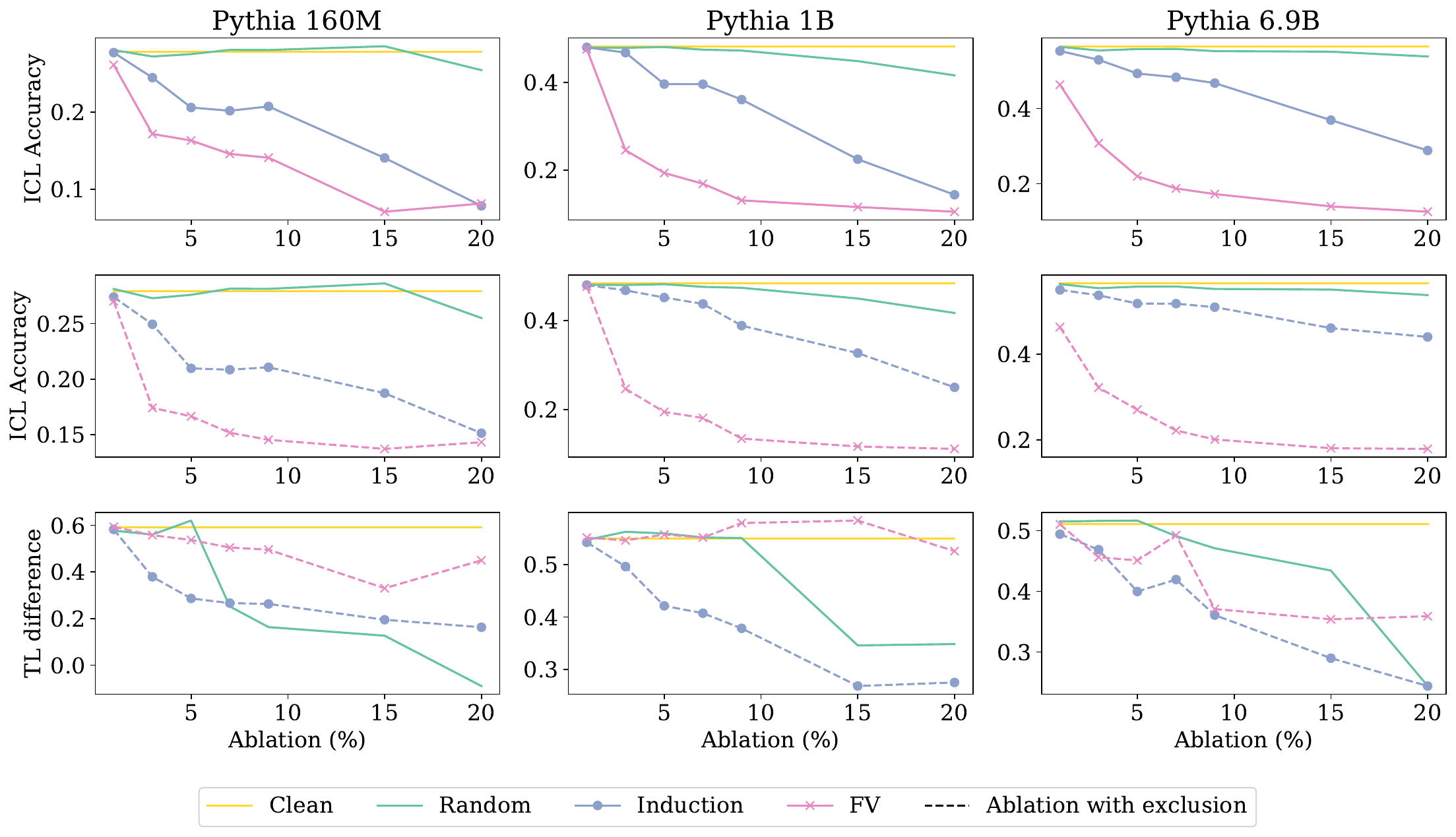}
    \vspace{-20pt}
  \caption{Top: Few-shot ICL accuracy after ablating induction and FV heads. Center: Few-shot ICL accuracy after ablating non-FV induction and non-induction FV heads. Bottom: Token-loss difference after ablating non-FV induction and non-induction FV heads. Ablating FV heads lead to a bigger drop in ICL accuracy, especially in larger models. Ablating induction heads with low FV scores does not significantly affect ICL accuracy. ICL accuracy and token-loss difference behave differently.}
  \label{fig:abl}
\end{figure*}

We evaluate the impact of ablating different proportions (1-20\%) of the top attention heads based on induction or FV score, across all models. We compare against two baselines: model performance with no ablation (``clean'') and with ablations of randomly sampled heads (``random'').  Figure \ref{fig:abl} shows results for three representative models, where few-shot ICL accuracy is averaged over the 19 evaluation tasks. We provide comprehensive results across all models and ICL accuracy broken down by task in Appendix \ref{app:abl_task}.

Our initial ablation experiments, shown in the top row of Figure \ref{fig:abl}, removed heads based on their scores without considering the potential overlap between induction and FV heads. These results revealed that ablating FV heads caused greater degradation in few-shot ICL performance compared to ablating induction heads, with this disparity becoming more pronounced in larger models. We also find that ablating induction heads has more effect on ICL performance than random. The effect of ablating induction heads converges to the effect of ablating FV heads as we increase the number of heads ablated.

However, the convergence noted above may be due to an increasing overlap in the set of heads ablated in the induction head and FV head ablations (Appendix \ref{app:overlap}). To address this, we conduct a second set of experiments using ablation with \textit{exclusion}, shown in the center row of Figure \ref{fig:abl}. When ablating induction heads while preserving the top 2\% FV heads, we observe minimal impact on few-shot ICL performance -- comparable to random ablations in models exceeding 1B parameters. Conversely, ablating FV heads while preserving induction heads continues to significantly impair ICL performance. The performance gap between FV and induction head ablations widens with model scale, suggesting that the earlier observed effects on induction head ablations were primarily driven by heads exhibiting both induction and FV properties.

These ablations suggest that the contributions of induction heads to ICL in the top row of Figure \ref{fig:abl} mostly come from heads that are both induction and FV heads, and that \textbf{FV heads matter the most for few-shot ICL}: as long as the model preserves its top 2\% FV heads, it can perform ICL with reasonable accuracy even if we ablate induction heads. 


The bottom row of Figure \ref{fig:abl} presents the effects of ablations with exclusion on token-loss difference. In smaller models (below 160M parameters), neither ablating induction nor FV heads shows significant impact to token-loss difference compared to random ablations. However, in models with over 345M parameters, induction head ablations affect token-loss difference more than FV head ablations, though this disparity decreases with model scale. This experiment primarily demonstrates that few-shot ICL accuracy and token-loss difference measure two very different things. These contrasting results between the two metrics help reconcile apparently contradictory findings in existing literature.



\section{FV heads evolve from induction heads}\label{sec:ckpt}

Finally, to further understand how these two families of attention heads develop, we analyze their evolution during model training. We examine attention heads across 8 intermediate training checkpoints in 7 Pythia models.

\subsection{Induction and FV strength during training}\label{sec:evolution}
\begin{figure*}
\centering
    \includegraphics[width=\linewidth]{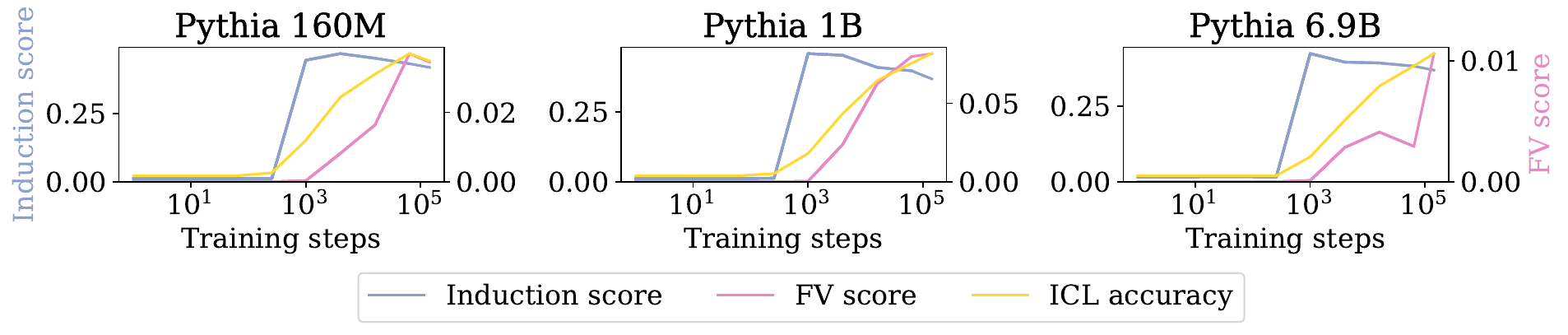}
  \caption{Evolution of induction and FV score averaged over top 2\% heads across training. Induction score rises sharply, then plateaus. FV score rises slightly later and gradually increases. ICL accuracy rises around the same time as induction and gradually increases.}
  \label{fig:mean_ckpt}
\end{figure*}

To measure the general strength of induction and FV mechanisms during training, we plot the mean induction and FV scores of the top 2\% induction and FV heads at each model checkpoint, along with few-shot ICL accuracy (Figure \ref{fig:mean_ckpt}). We include plots for all Pythia models in Appendix \ref{app:train_scores}.

Our analysis reveals a consistent pattern across all Pythia models: induction heads emerge early in training, at around step 1,000 out of 143,000, while FV heads appear substantially later at around step 16,000. The development of these heads shows distinct characteristics as well -- induction scores exhibit a sharp initial rise followed by a plateau or slight decline, whereas FV scores demonstrate a gradual but sustained increase from step 16,000 through the end of training. This temporal asymmetry suggests that induction heads represent a simpler mechanism that models can acquire earlier, while FV heads embody a more complex mechanism that requires extended training. In addition, we observe that in all models, few-shot ICL accuracy begins to improve around the same time as when induction heads appear, and continues to gradually increase throughout training.

\subsection{Evolution of individual heads during training}
\begin{figure*}
\centering
    \includegraphics[width=0.8\linewidth]{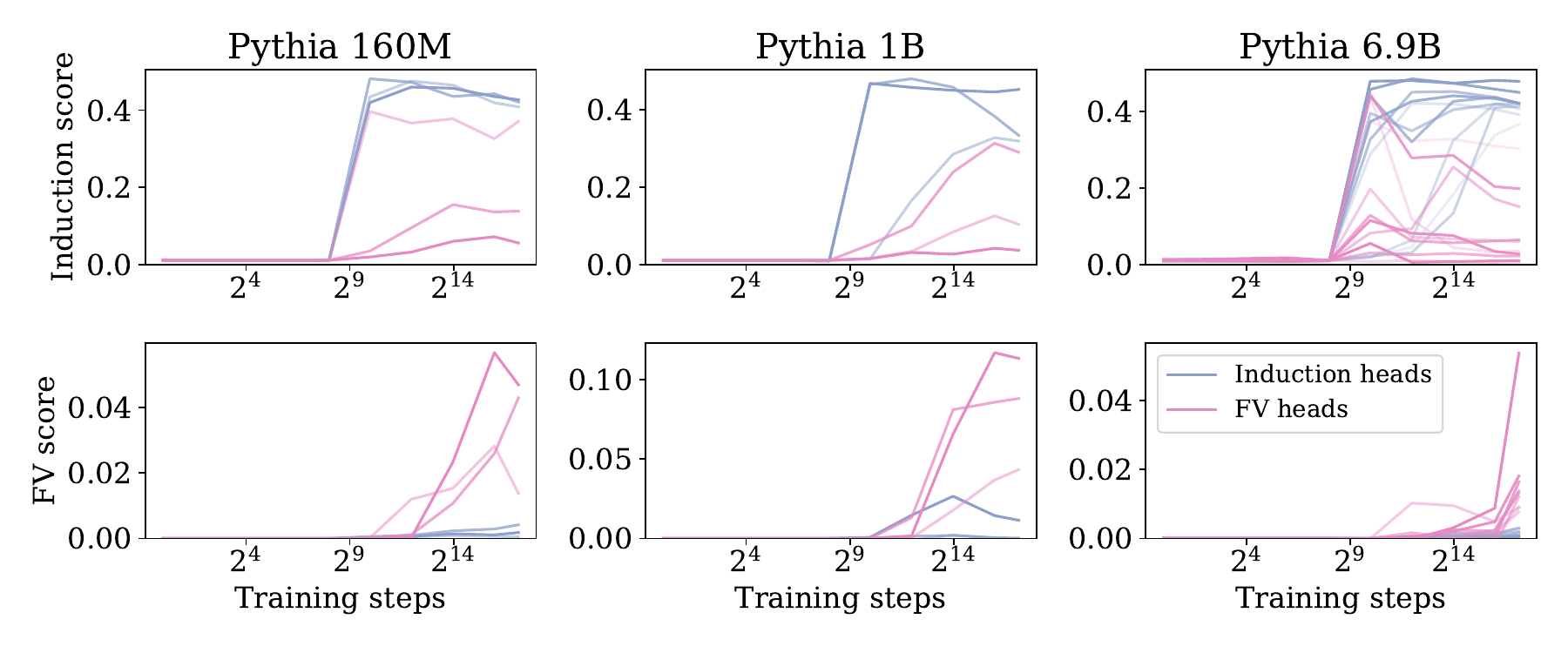}
  \caption{Evolution of induction scores (top) and FV scores (bottom) of individual induction and FV heads across training. Certain FV heads have a high induction score earlier in training; the reverse is not true for induction heads.}
  \label{fig:indv_ckpt}
\end{figure*}

To gain more granular insights into head development, we investigate the evolution of individual attention heads throughout training. Figure \ref{fig:indv_ckpt} the induction scores (top row) and FV scores (bottom row) of the top 2\% induction and FV heads across training steps. Individual heads are represented by continuous lines, with line opacity corresponding to their final induction or FV scores.

A striking pattern emerges across all models: many heads that ultimately become strong FV heads initially exhibit high induction scores, emerging around the same time as dedicated induction heads. These proto-FV heads initially achieve induction scores comparable to those of specialized induction heads. However, as training progresses, their induction scores gradually decline while their FV scores increase. Importantly, this pattern is unidirectional; we found no instances of induction heads that develop significant FV capabilities during training, as evidenced by their consistently low FV scores throughout training. This suggests \textbf{many FV heads evolve from induction heads during training}, but not vice versa.


\section{Interpretation and discussion}\label{sec:disc}


Our investigation revealed several key insights about the relationship between induction and FV heads in transformer models and their effect on ICL. While these mechanisms are distinct, they show notable correlation (\S\ref{sec:overlap}). FV heads consistently appear in deeper layers and emerge later in training compared to induction heads (\S\ref{sec:overlap},\ref{sec:evolution}). Our ablation studies demonstrated that FV heads are crucial for few-shot ICL performance, particularly in larger models, while induction heads have comparatively minimal impact (\S\ref{sec:abl}). Furthermore, we observed multiple instances of heads transitioning from induction to FV functionality during training, but never the reverse (\S\ref{sec:evolution}). We propose two working conjectures to explain these empirical findings more broadly, and consider arguments for and against them.



Our first conjecture (C1) posits that \textbf{induction heads are an early version of FV heads}. Under this interpretation, induction heads serve as a stepping stone for models to develop the more sophisticated FV mechanism. As FV heads emerge and prove more effective at ICL tasks, they gradually supersede the simpler induction mechanism.

Several lines of evidence support this conjecture. First, we observed multiple FV heads that initially displayed strong induction behavior before transitioning to FV functionality. The subsequent decline in induction scores suggests these heads abandon the simpler mechanism once the more effective FV capability develops. The unidirectional nature of this transition - we never observe induction heads with initially high FV scores - further supports this interpretation. Additionally, the minimal effect of ablating pure induction heads (those with low FV scores) on few-shot ICL performance suggests their role becomes less critical once FV heads develop. To further verify this, future work could explore how removing induction heads during training could impact the development of FV heads. However, C1 does not fully explain the existence of FV heads with low induction scores throughout training.

C1 also aligns with our observations about architectural and training dynamics. FV heads consistently emerge later in training and appear in deeper layers, consistent with them implementing a more complex computation. In addition, ICL performance begins to increase at the same time as the emergence of induction heads, but continues to improve after induction heads form, similarly to how FV heads gradually increase. This may indicate that the sharp emergence of induction heads contributes to an initial rise in ICL performance, but the emergence of the FV mechanisms contributes to further improvements in ICL.



An alternative conjecture (C2) suggests that \textbf{FV heads are a combination of induction and another mechanism}. This conjecture proposes that heads appearing to ``transition'' from induction to FV functionality are actually polysemantic heads that implement both, and possibly other mechanisms. Their measured induction scores decline as their attention patterns diversify to support multiple mechanisms.


While C2 explains the correlation between induction and FV mechanisms as arising from shared underlying mechanisms, it faces a significant challenge: our ablation studies show that removing monosemantic FV heads (those without significant induction scores) substantially hurts ICL performance. This is difficult to reconcile with C2's prediction that pure FV heads should be less critical if the key functionality depends on combined induction-FV mechanisms.



\section{Conclusion}

Our research challenges the prevailing understanding of in-context learning mechanisms in transformer models. While induction heads have been widely considered the primary driver of ICL, our evidence demonstrates that function vector (FV) heads play a more crucial causal role in few-shot ICL performance. We attribute previous misconceptions to two key factors: the conflation of few-shot ICL with token-loss difference metrics, as well as not accounting for the overlap between induction and FV heads.

Remarkably, although induction and FV mechanisms appear to implement two distinct processes, we also observe an interesting interplay between the two types of heads: induction and FV scores are correlated, and many FV heads are ``former'' induction heads with high induction scores earlier in training. This observation supports an early conjecture where induction heads serve as precursors to FV heads: the simpler induction mechanism provides an initial foundation for ICL capability, from which the more sophisticated FV mechanism eventually emerges.

Our investigation also yields important methodological insights for the broader field of model interpretability. First, seemingly equivalent definitions of model capabilities (such as few-shot accuracy versus token-loss difference) can lead to substantially different conclusions. Second, studying mechanistic components in isolation may produce misleading results when these components share overlapping behaviors, as demonstrated by the confounding effects of ablating heads that exhibit both high induction and FV scores.

Furthermore, our results challenge strong versions of the universality hypothesis in interpretability. While both induction and FV heads contribute meaningfully to few-shot ICL in smaller models, their relative importance diverges with scale -- FV heads become increasingly crucial while induction heads' impact approaches that of random ablations. This scale-dependent behavior suggests that mechanisms may vary across model architectures.

These findings prompt several important questions for future research. If induction heads indeed serve as precursors to FV heads, what makes this necessary? What role do the remaining induction heads serve in fully trained models? 
Finally, are there additional mechanisms that could provide an even more complete explanation of ICL capabilities? 

\section*{Acknowledgements}
We thank Jiahai Feng, Neel Nanda, Robert Kirk, and Anish Kachinthaya for their helpful feedback. KY is supported by the Vitalik Buterin Ph.D. Fellowship in AI Existential Safety.

\bibliographystyle{Styles/icml2025}
\bibliography{main}

\appendix

\section{Appendix}
\subsection{Induction scores vs. FV scores}\label{app:corr}
\begin{figure*}
\centering
    \includegraphics[width=\linewidth]{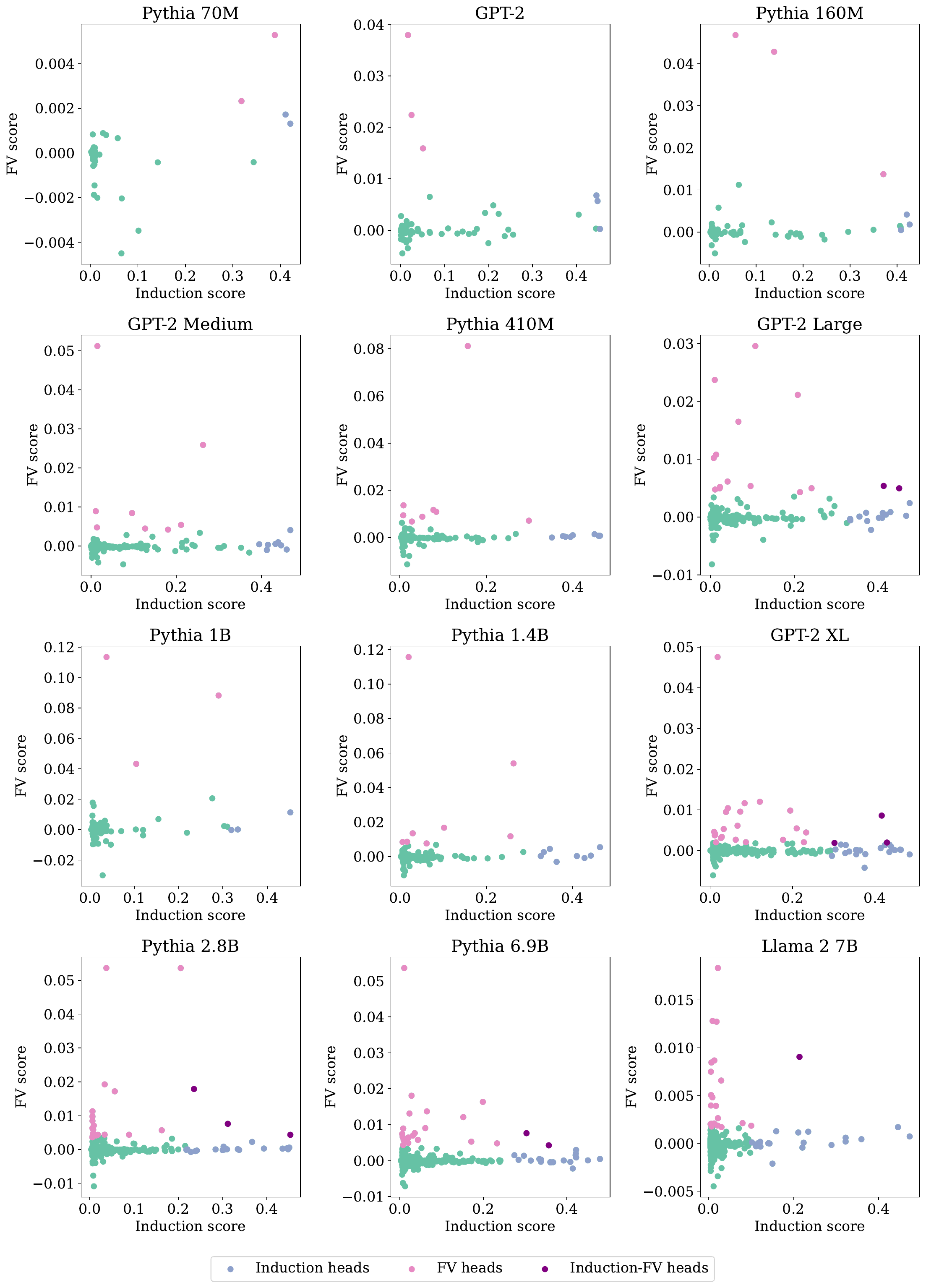}
  \caption{Induction and FV scores of attention heads.}
  \label{fig:corr}
\end{figure*}

In Figure \ref{fig:corr}, we plot the induction score and FV score of each attention head. 

\subsection{Ablations}\label{app:abl}
\begin{figure*}
\centering
    \includegraphics[width=\linewidth]{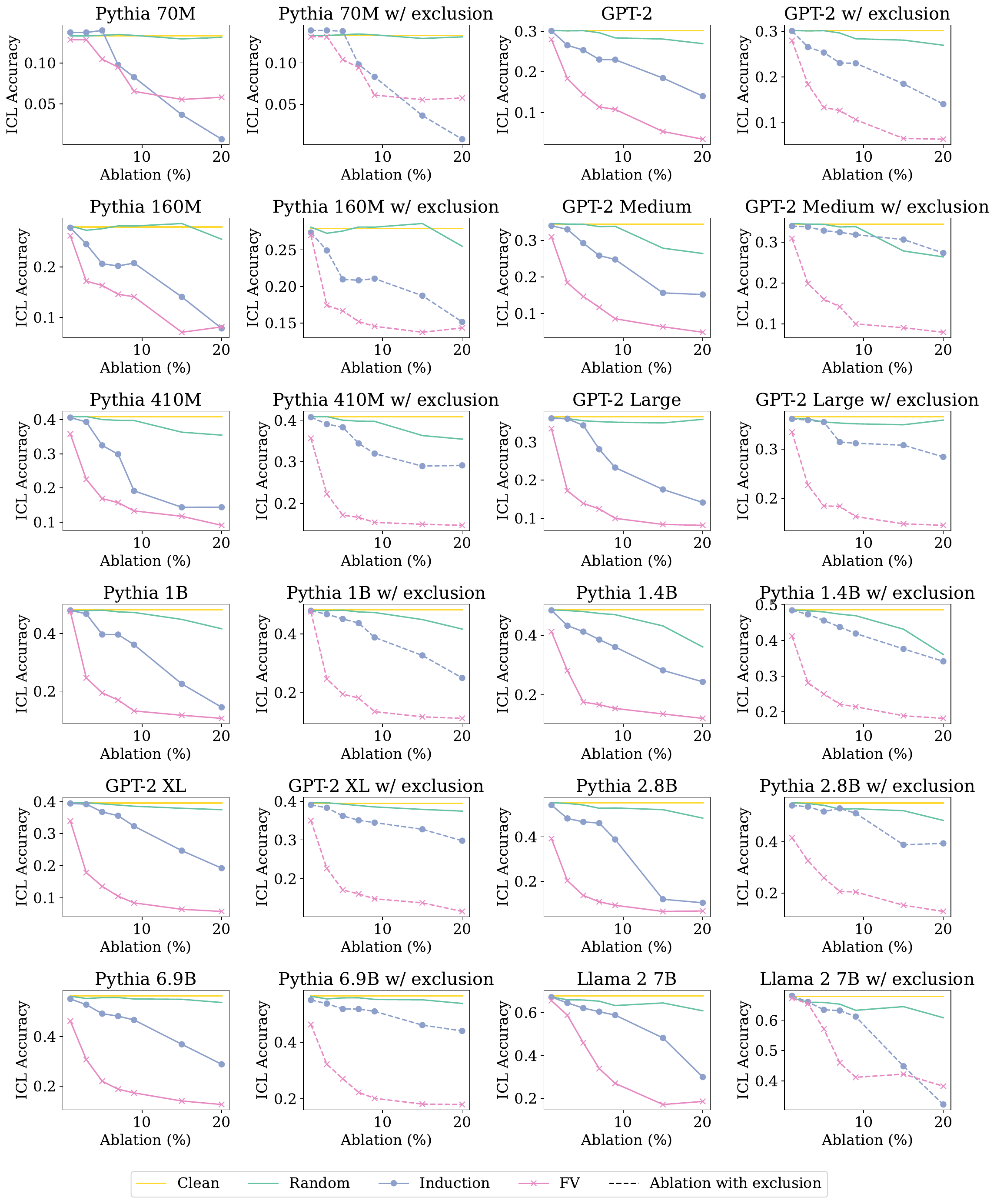}
  \caption{ICL accuracy after ablating induction and FV heads.}
  \label{fig:abl_all}
\end{figure*}

\begin{figure*}
\centering
    \includegraphics[width=\linewidth]{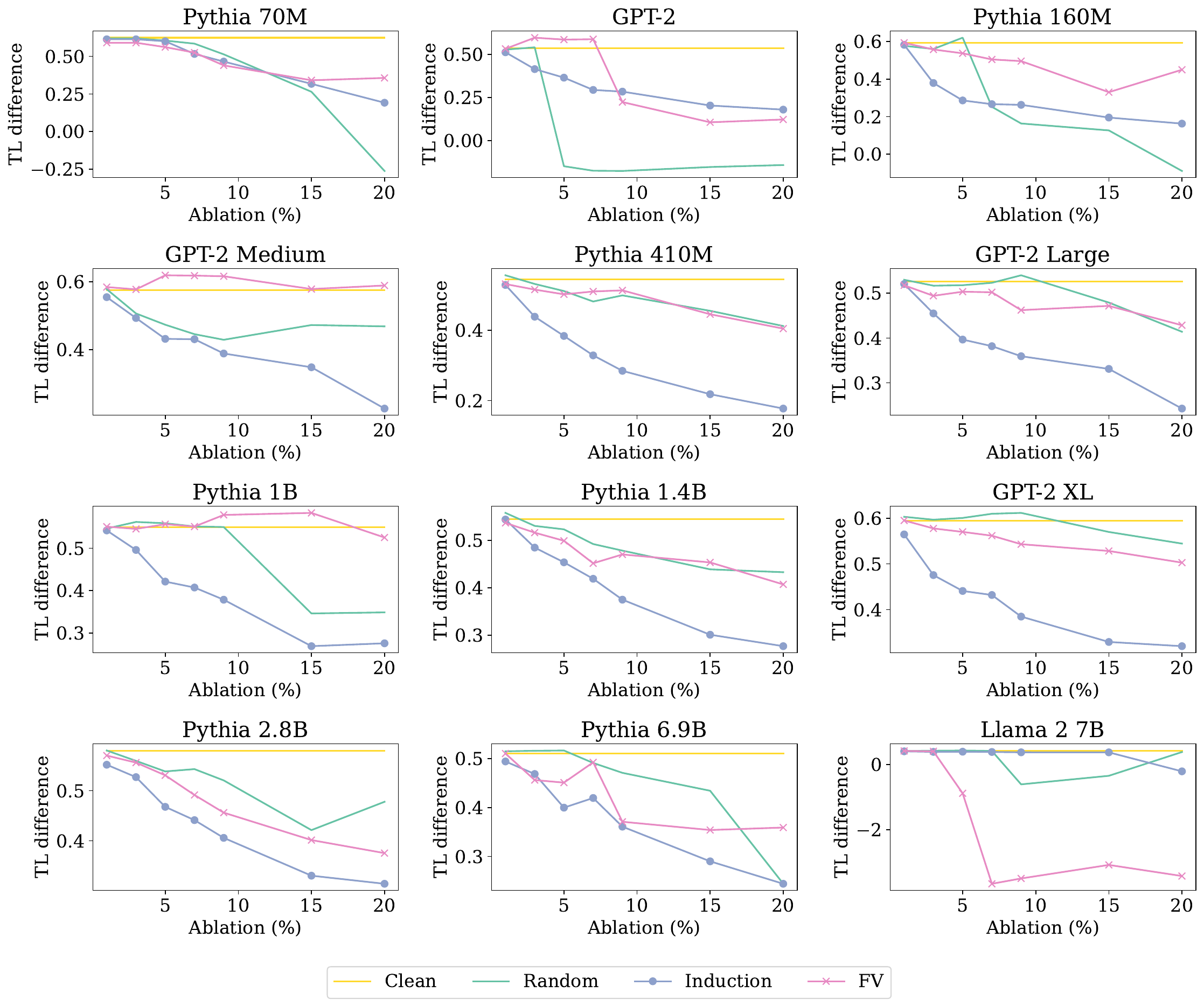}
  \caption{Token-loss difference after ablating induction heads with low FV scores and FV heads with low induction scores.}
  \label{fig:icl_scores_all}
\end{figure*}

In Figure \ref{fig:abl_all}, we plot model accuracy averaged over ICL tasks across different quantities of heads ablated in each head type. In Figure \ref{fig:icl_scores_all}, we plot the token-loss difference of models across different quantities of heads ablated.

\subsection{Random and zero ablations}\label{app:zero_ablate}
\begin{figure*}
\centering
    \includegraphics[width=\linewidth]{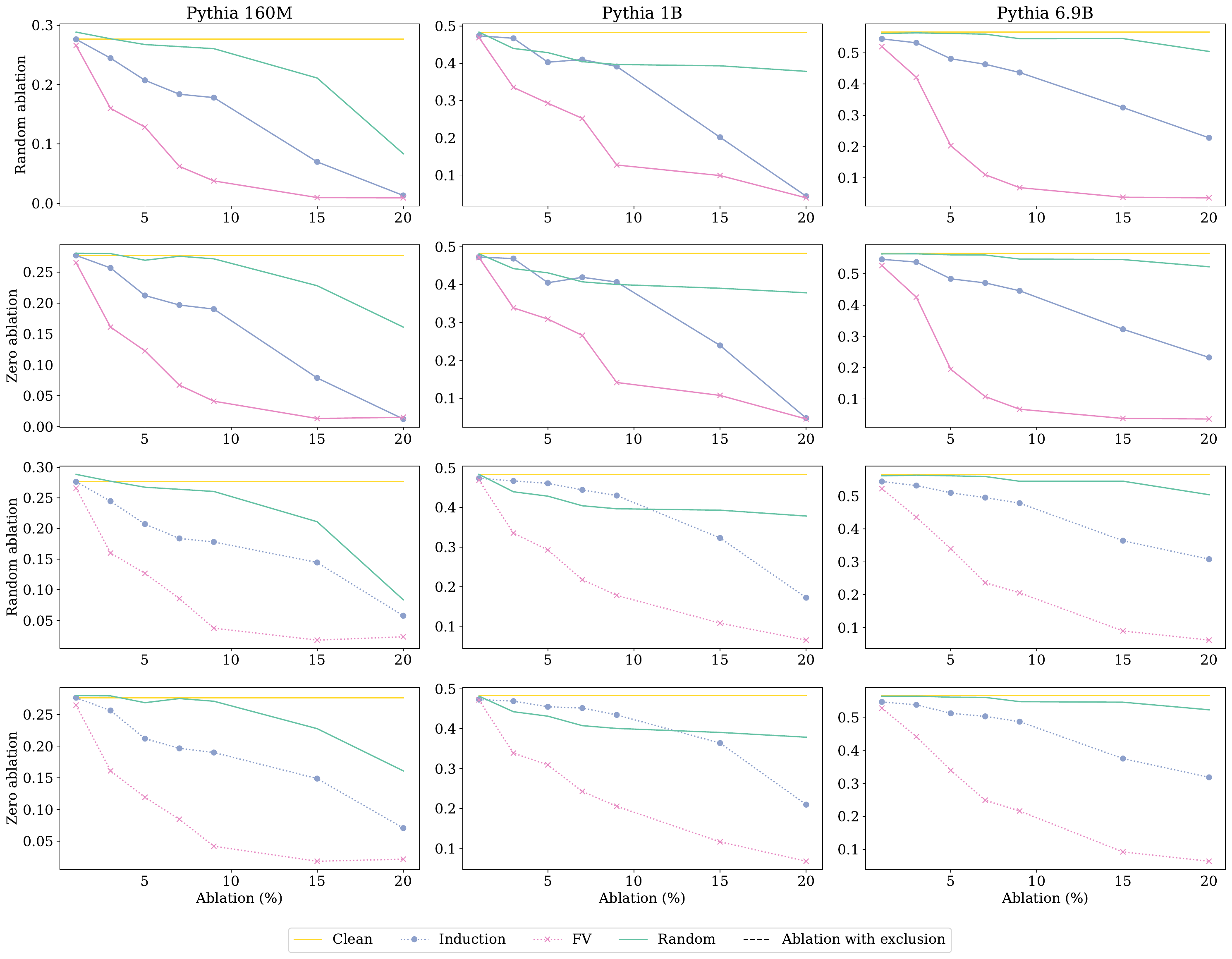}
  \caption{ICL accuracy after ablating induction heads and FV heads with random or zero ablation.}
  \label{fig:random_zero_abl}
\end{figure*}

In Figure \ref{fig:random_zero_abl}, we plot model accuracy averaged over ICL tasks across different quantities of heads ablated with random ablation or zero ablation. For random ablations, we replace the head's output vector with the output vector of a randomly sampled different head. For zero ablations. we replace the head's output vector with a zero vector.

\subsection{Ablating random heads at specific layers}
\begin{figure*}
\centering
    \includegraphics[width=\linewidth]{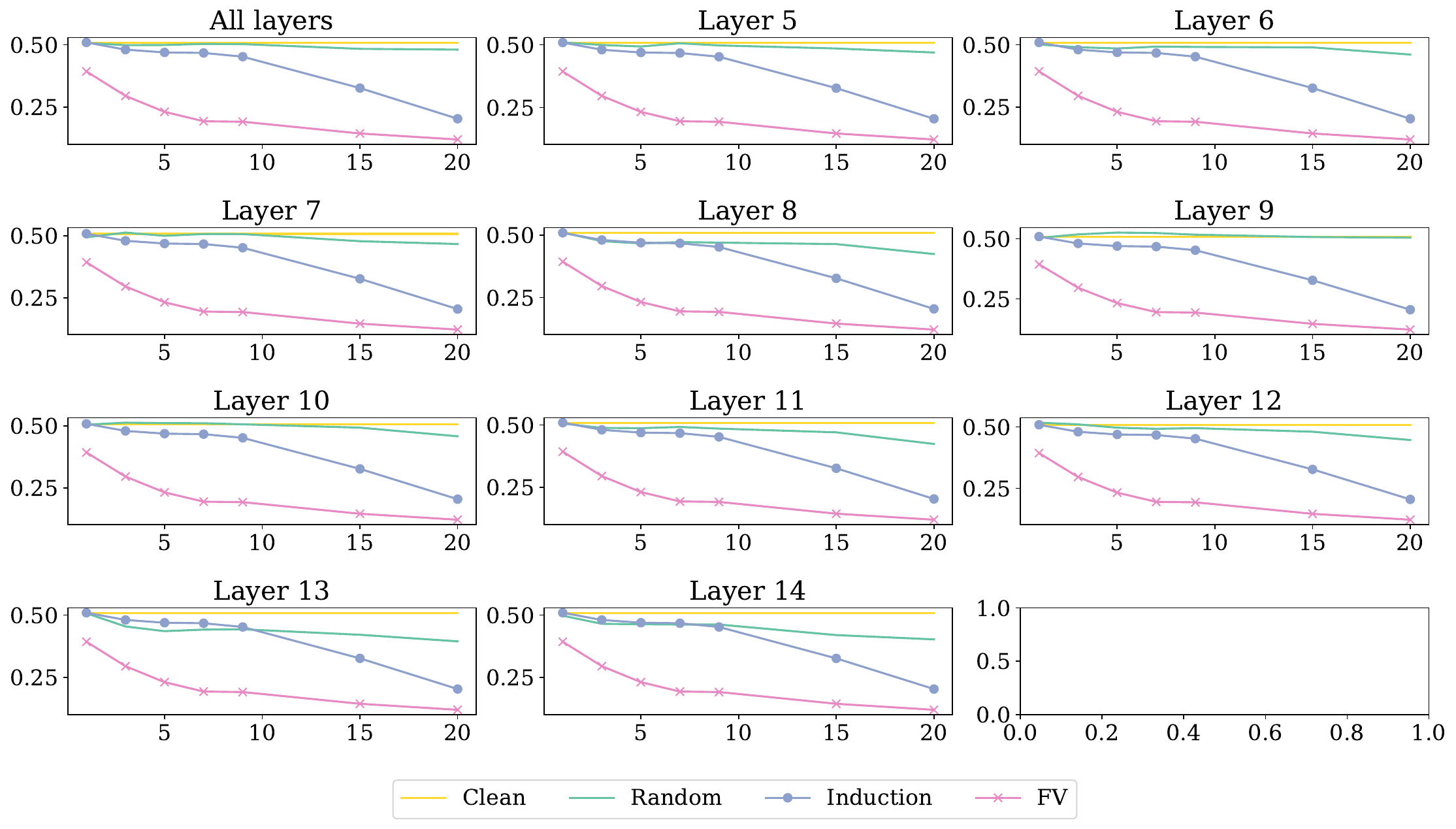}
  \caption{ICL accuracy after ablating randomly sampled heads from specific layers. The clean ICL accuracy, induction ablations and FV ablations are also plotted for comparison but only the random ablations (green curve) are affected by the choice of target layer.}
  \label{fig:abl_rand}
\end{figure*}

In Figure \ref{fig:abl_rand}, we ablate heads randomly sampled from specific layers of the model. Let $L$ be the number of heads in each layer, $A$ be the number of heads we're ablating, and $\ell$ be the layer we're targeting. Then, if $A < L$, we sample $A$ heads from layer $\ell$. If $A \geq L$, we ablate all $L$ heads in layer $\ell$ and we sample $A - L$ heads from other layers to ablate.

\subsection{Induction and function vector scores across models}\label{sec:scores}

\begin{figure*}
\centering
    \includegraphics[width=0.9\linewidth]{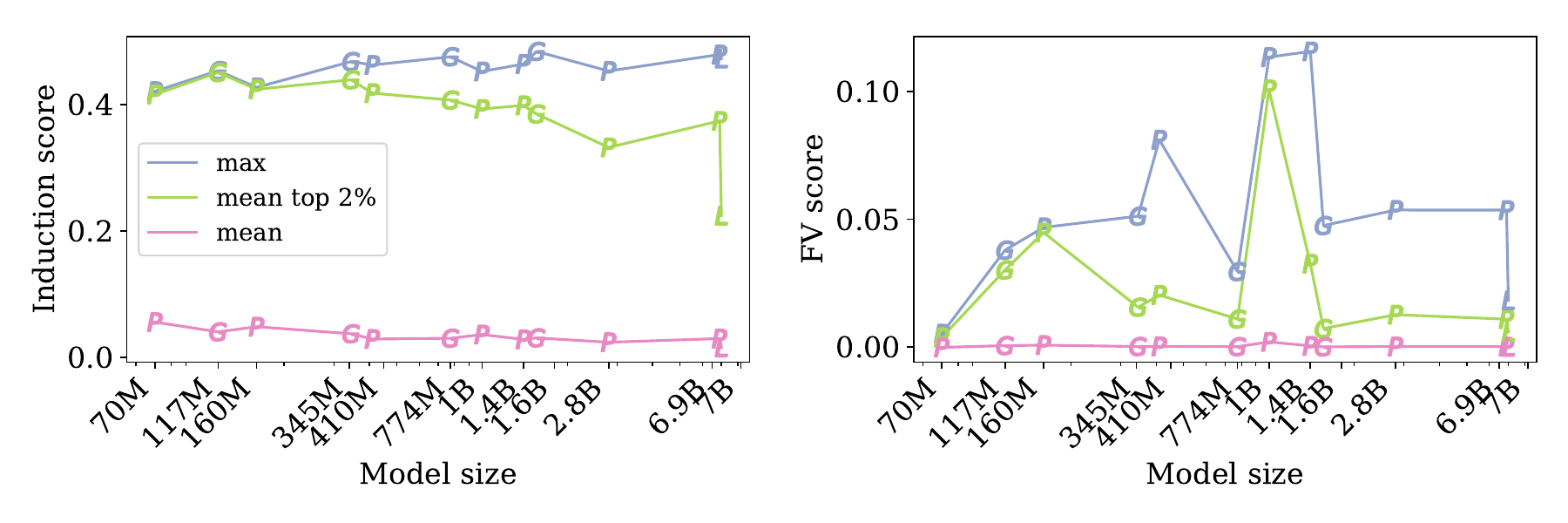}
  \caption{Induction score (left) and FV score (right) of attention heads across model size. We plot the maximum score of all heads, mean of the top 2\% scores, and mean score of all heads. Overall, induction scores are similar across models. Pythia 70M and Llama 2 have relatively low FV scores, Pythia 1B and 1.4B have relatively high FV scores.}
  \label{fig:head_strength}
\end{figure*}

\begin{figure}[h]
    \centering
    \begin{subfigure}{\linewidth}
        \centering
        \includegraphics[width=\linewidth]{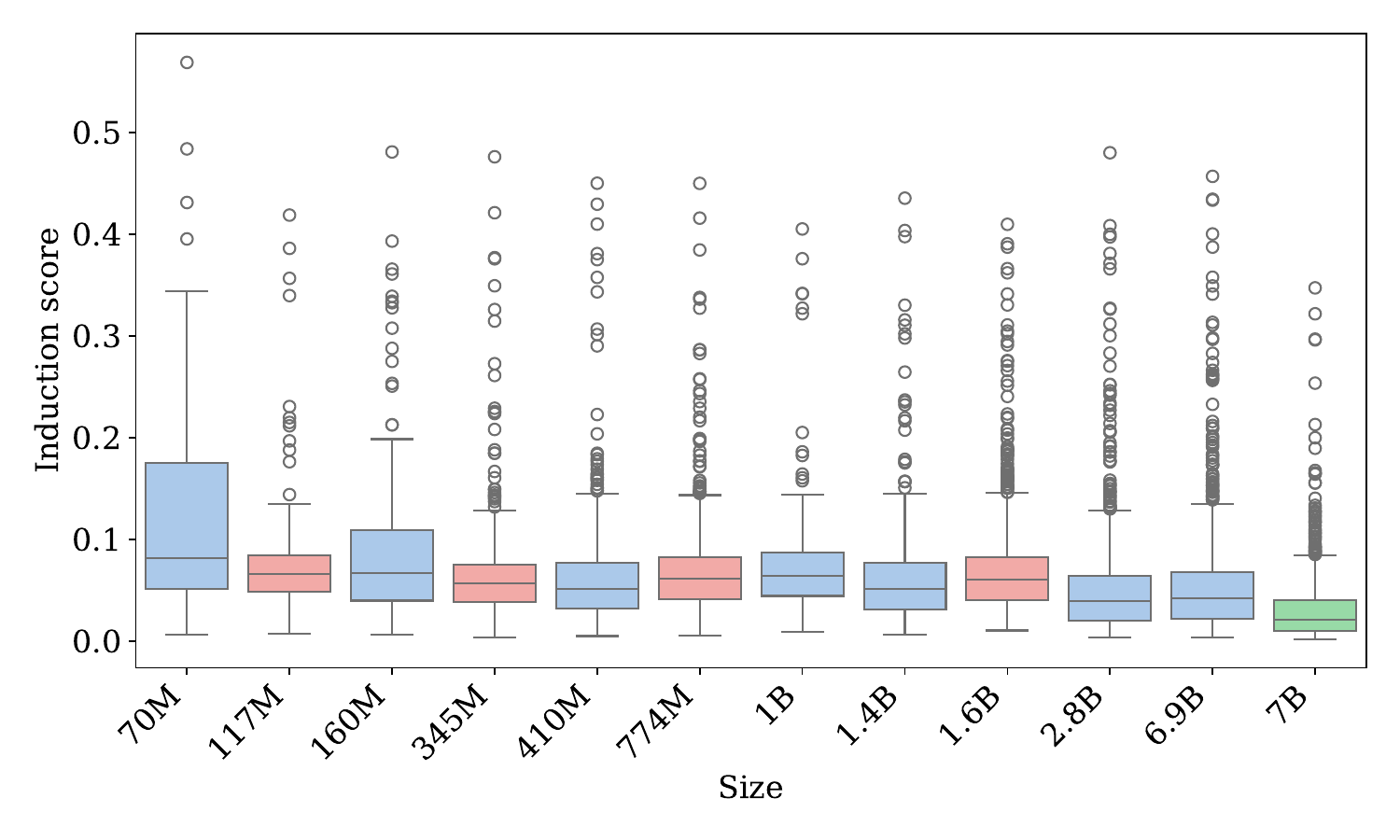} 
    \end{subfigure}
    \vspace{0.5cm} 
    \begin{subfigure}{\linewidth}
        \centering
        \includegraphics[width=\linewidth]{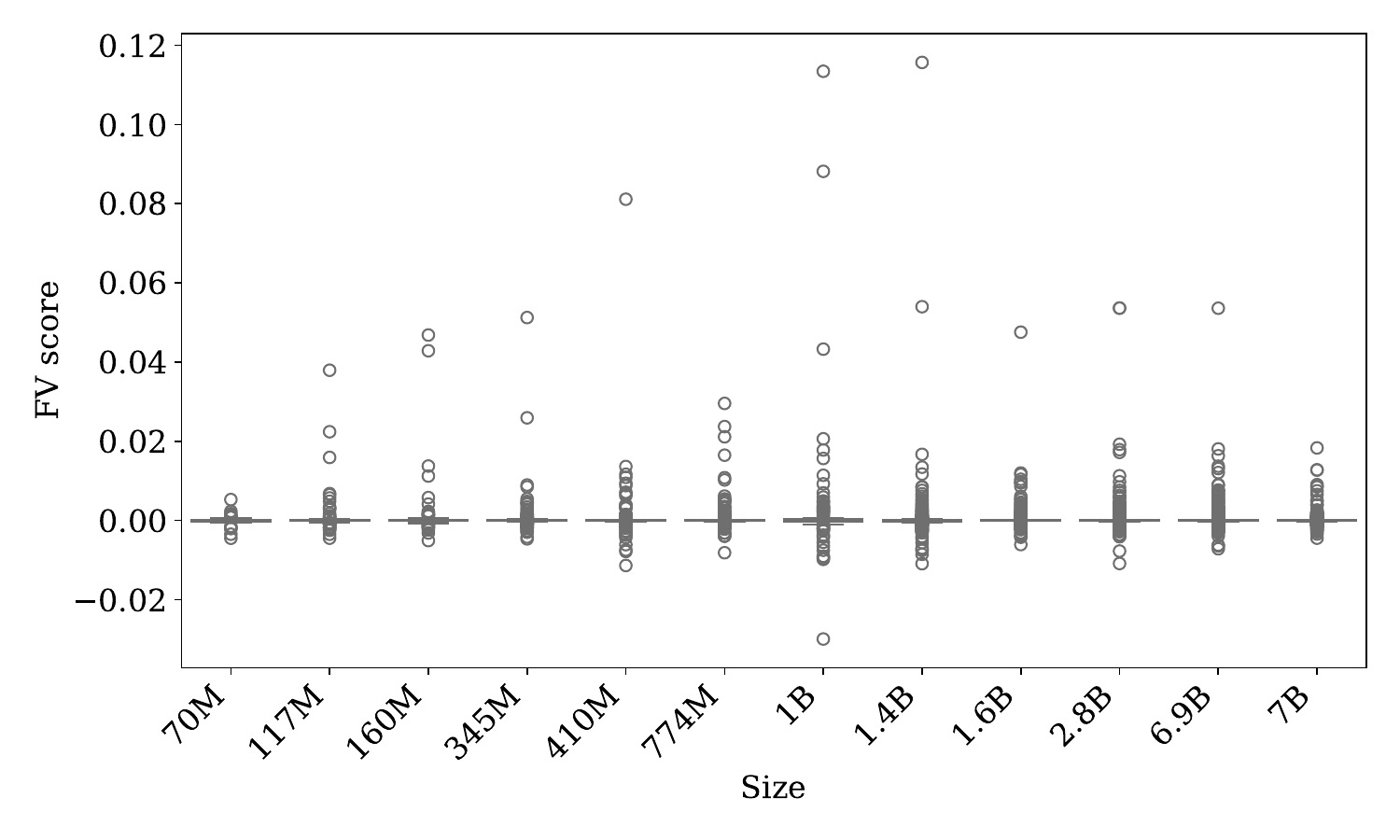} 
    \end{subfigure}
    \caption{Distribution of induction scores (top) and FV scores (bottom) across model size.}
    \label{fig:box}
\end{figure}

Our ablation studies reveal a consistent trend where FV heads are increasingly important relative to induction heads for ICL performance as model scale increases. To further explore this trend, we examine how induction scores and FV scores vary with model scale, and whether these scores follow similar trends to our ablation experiments.

In Figure \ref{fig:head_strength}, we plot the maximum and mean induction and FV scores across all heads, and mean scores of top 2\% heads, for each model. The left plot in Figure \ref{fig:head_strength} shows that induction scores are relatively similar across model size, with a small increase in maximum induction score and a decrease in the top 2\% mean induction score with model scale.

In the right plot of Figure \ref{fig:head_strength}, there is no clear trend between FV score and model scale, however, Pythia 1B and 1.4B models have markedly higher maximum FV scores. One possible explanation is that models with high head dimensionality relative to total parameter count have stronger FV heads: Pythia 1B and 1.4B have head dimensionality of 256 and 128 respectively (Table \ref{tab:models}) whereas other models with similar parameter count have only 64-80 attention head dimensions. 

We also find very low FV scores in Pythia 70M and Llama 2 models. FV scores may be low in Pythia 70M because it is too small in parameter size for FV heads to emerge. Low scores in Llama 2 compared to other models may be due to differences in architecture, and additional experiments can help confirm this. Overall, we do not recover the same trend in induction/FV scores as the trend in our ablation studies. 

For reference, we also provide box plots of the full distribution of induction and FV scores in Figure \ref{fig:box}. 

\subsection{Evaluating function vectors on task execution}\label{sec:fvtask}

To further inspect the prevalence of the FV mechanism in different models, we evaluate the efficacy of FVs for ICL task execution. A successful FV triggers the model to execute the particular task the FV encodes, even when the model sees no useful in-context demonstrations of the task. 
First, to extract FVs, for each model we gather the top 2\% attention heads with highest FV scores as the set $\mathcal{A}$. Then, for each ICL task $t\in \mathcal{T}$, we sum the average outputs of heads in $\mathcal{A}$ over prompts from $t$ and obtain the FV for the task $t$:
$
FV_t = \sum_{a\in\mathcal{A}} \bar{a}^t.
$

\begin{figure}
\centering
    \includegraphics[width=\linewidth]{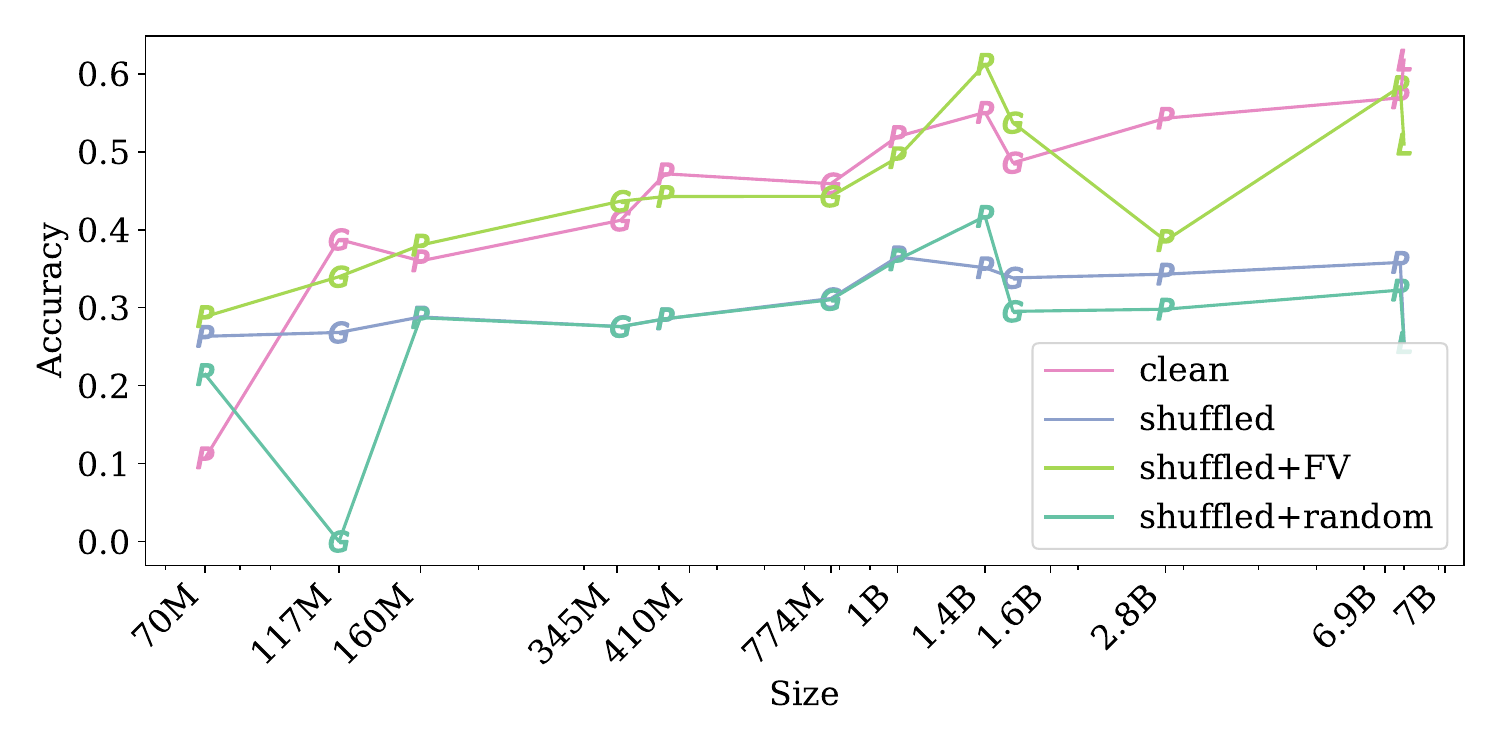}
  \caption{Model ICL accuracy on prompts with 10 in-context examples (clean), on uninformative shuffled prompts, on shuffled prompts with FV, and on shuffled prompts with random head outputs. Adding FV recovers most of the model accuracy on a clean run, with the exception of Pythia 2.8B.}
  \label{fig:fv_intervention}
\end{figure}

In Figure \ref{fig:fv_intervention}, we report model accuracy averaged over 40 ICL tasks where the model performs inference on uncorrupted prompts (clean), prompts with shuffled labels (shuffled), shuffled prompts with $FV_t$ added to hidden states at layer $|L|/3$, and shuffled prompts with FV extracted from random heads added to hidden states at layer $|L|/3$. We take 1000 examples per task that are previously unseen during FV score computation.

In most models, adding the FV recovers model performance on uncorrupted prompts, with the exception of Pythia 2.8B. One possible explanation for this is again due to head dimensionality: Pythia 2.8B has head dimension 80, which is significantly smaller than other models with similar parameter size that have head dimensions of 128. Together with our experiments in \S\ref{sec:scores}, results provide preliminary evidence that \textbf{high head dimensionality relative to model size is a predictor of FV strength (H6)}.

\subsection{ICL tasks}\label{app:tasks}
In Table \ref{tab:task_list}, we list the ICL tasks used in this study. We refer to \citet{fv} and \citet{binding} for a detailed description of each task.
\begin{table*}[ht]
    \centering
        \caption{Summary of ICL tasks used in our study. Tasks in \textbf{bold} are new tasks that were not used in \citet{fv}.}
    \resizebox{0.6\linewidth}{!}{\begin{tabular}{ll}
        \toprule
        \textbf{Task Name} & \textbf{Task Source} \\
        \midrule
        \multicolumn{2}{l}{\textbf{Abstractive Tasks}} \\
        \midrule
        \textbf{Abstract clf }&  \\
        Antonym & \citet{nguyen-etal-2017-distinguishing} \\
        \textbf{Binding capital} & \citet{binding}\\
        \textbf{Binding capital parallel} & \citet{binding}\\
        \textbf{Binding fruit} & \citet{binding}\\
        \textbf{Binding shape} & \citet{binding}\\
        Capitalize first letter & \citet{nguyen-etal-2017-distinguishing} \\
        \textbf{Capitalize index} & \\
        \textbf{Capitalize second letter} & \\
        Capitalize & \citet{nguyen-etal-2017-distinguishing} \\
        Country-capital & \citet{fv}\\
        Country-currency & \citet{fv}\\
        English-French & \citet{conneau2017word} \\
        English-German & \citet{conneau2017word} \\
        English-Spanish & \citet{conneau2017word} \\
        \textbf{French-English} & \citet{conneau2017word} \\
        Landmark-Country & \citet{hernandez2024linearity} \\
        Lowercase first letter & \citet{fv}\\
        National parks & \citet{fv}\\
        Next-item & \citet{fv}\\
        Previous-item & \citet{fv}\\
        Park-country &\citet{fv} \\
        Person-instrument & \citet{hernandez2024linearity} \\
        Person-occupation & \citet{hernandez2024linearity} \\
        Person-sport & \citet{hernandez2024linearity} \\
        Present-past & \citet{fv}\\
        Product-company & \citet{hernandez2024linearity} \\
        Singular-plural & \citet{fv}\\
        Synonym & \citet{nguyen-etal-2017-distinguishing} \\
        \midrule
        CommonsenseQA (MC-QA) & \citet{talmor-etal-2019-commonsenseqa} \\
        Sentiment analysis (SST-2) & \citet{socher-etal-2013-recursive} \\
        AG News & \citet{zhang2015character} \\
        \midrule
        \multicolumn{2}{l}{\textbf{Extractive Tasks}} \\
        \midrule
        Adjective vs. verb & \citet{fv}\\
        Animal vs. object & \citet{fv}\\
        Choose first of list & \citet{fv}\\
        Choose middle of list & \citet{fv}\\
        Choose last of list & \citet{fv}\\
        Color vs. animal & \citet{fv}\\
        Concept vs. object & \citet{fv}\\
        Fruit vs. animal & \citet{fv}\\
        Object vs. concept & \citet{fv}\\
        Verb vs. adjective & \citet{fv}\\
        \midrule
        CoNLL-2003, NER-person & \citet{tjong-kim-sang-de-meulder-2003-introduction} \\
        CoNLL-2003, NER-location & \citet{tjong-kim-sang-de-meulder-2003-introduction} \\
        CoNLL-2003, NER-organization & \citet{tjong-kim-sang-de-meulder-2003-introduction} \\
        \bottomrule
    \end{tabular}}
    \label{tab:task_list}
\end{table*}

\subsection{Ablations by task}\label{app:abl_task}

\begin{figure*}[h]
    \centering
    \begin{subfigure}{\linewidth}
        \centering
        \includegraphics[width=0.8\linewidth]{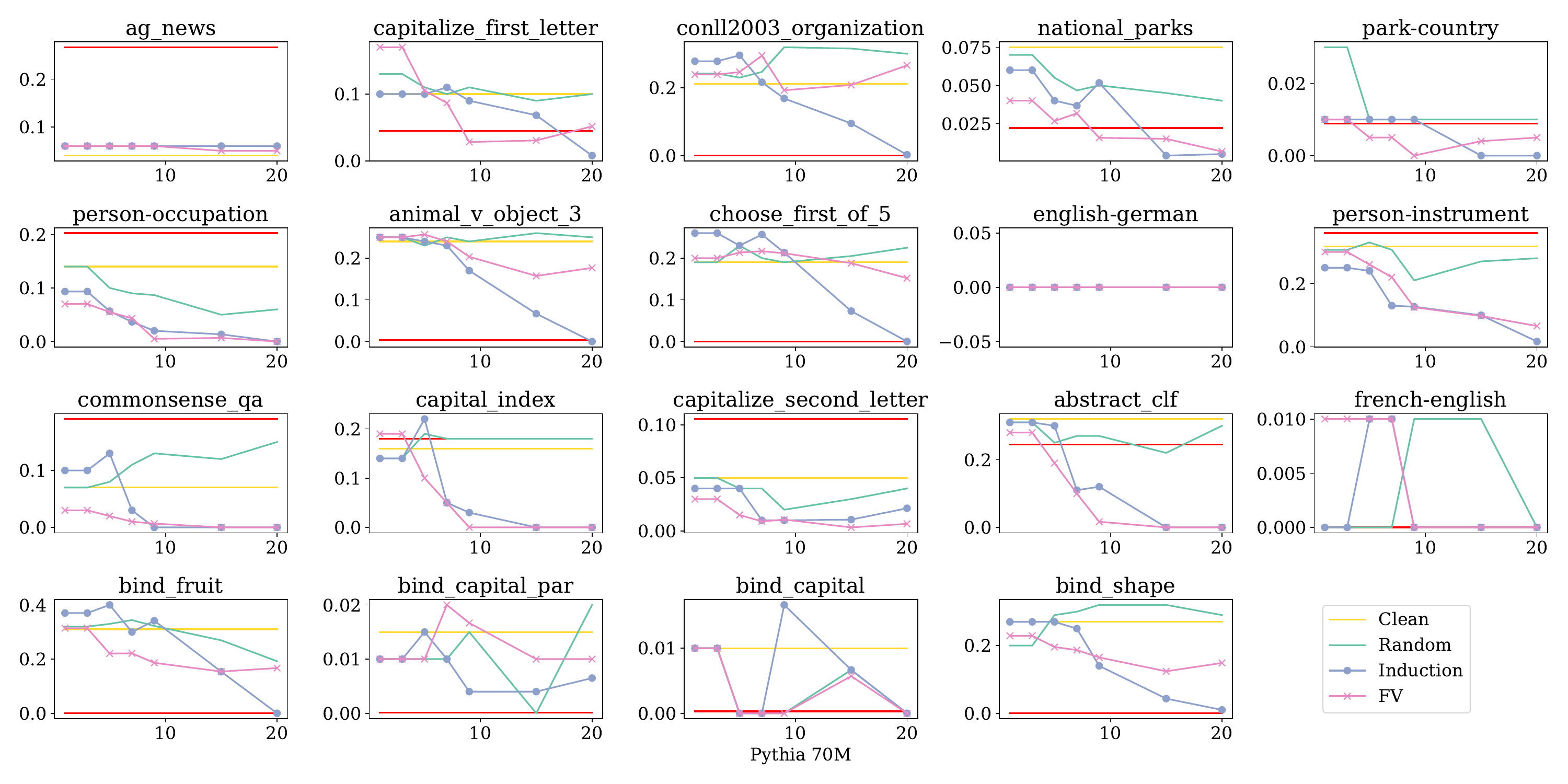} 
    \end{subfigure}
    \vspace{0.5cm}
\begin{subfigure}{\linewidth}
        \centering
        \includegraphics[width=0.8\linewidth]{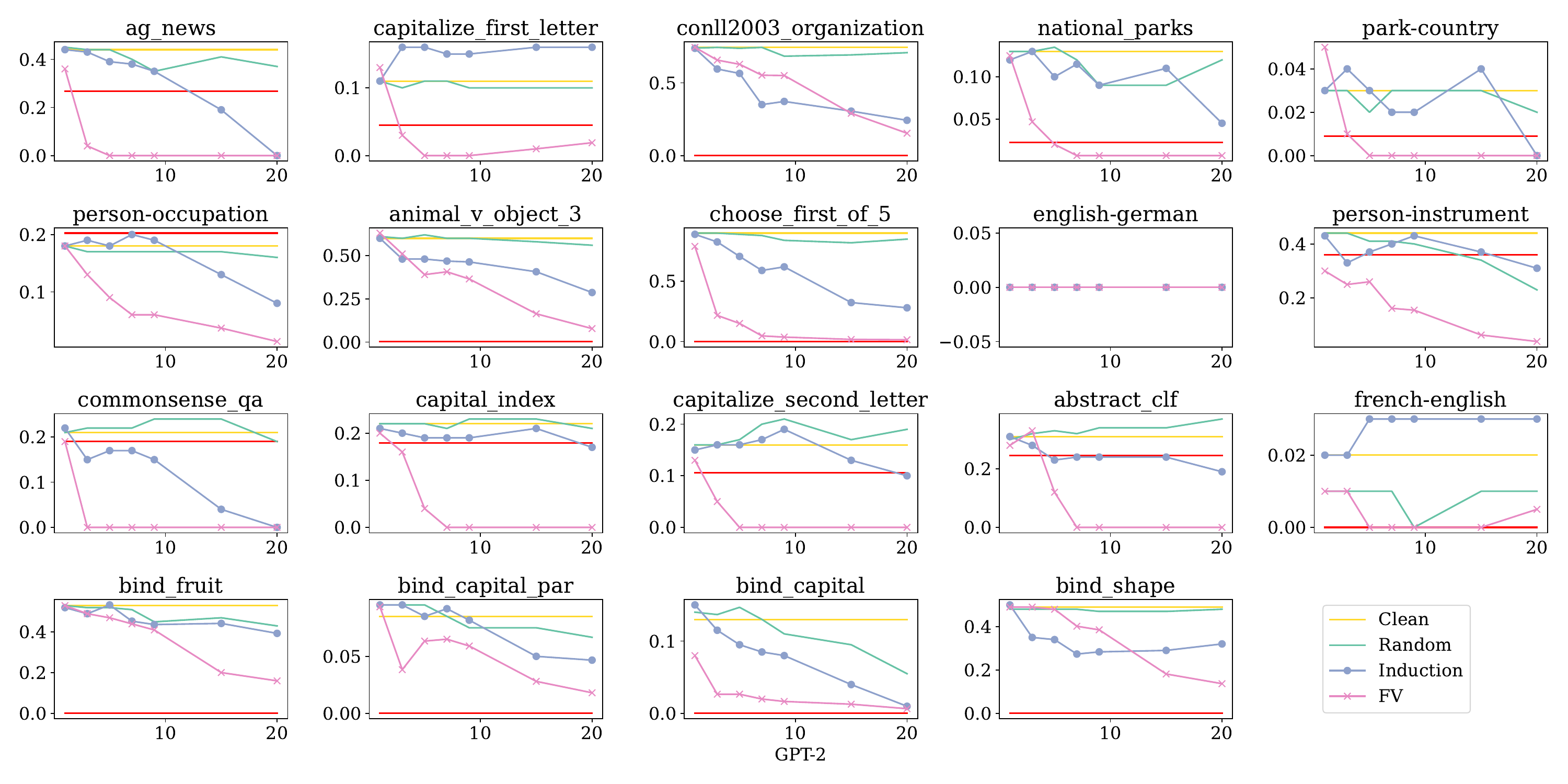} 
    \end{subfigure}
    \vspace{0.5cm}
\begin{subfigure}{\linewidth}
        \centering
        \includegraphics[width=0.8\linewidth]{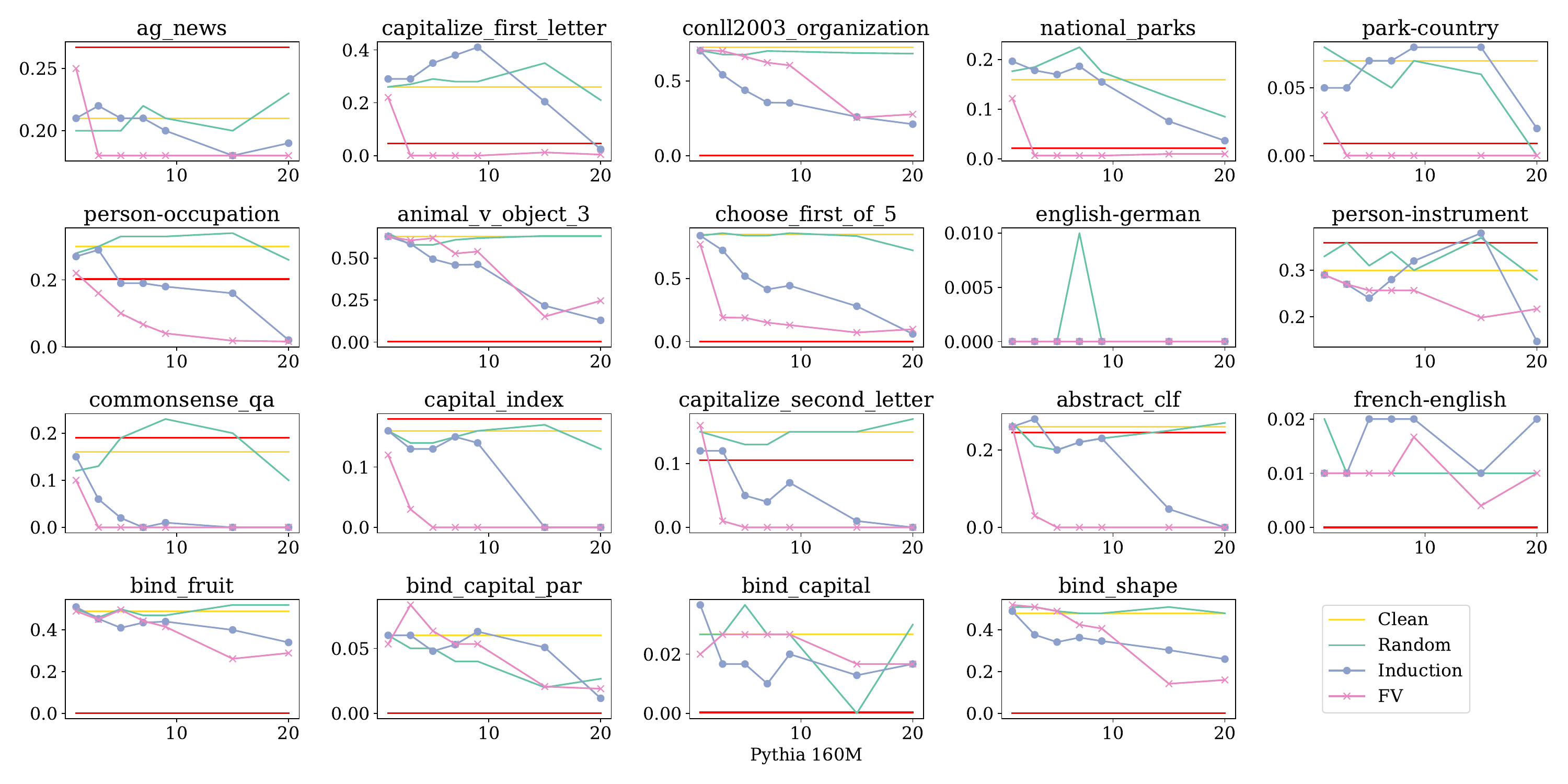} 
    \end{subfigure}
        \caption{ICL accuracy after ablations by task. The red horizontal line represents the random baseline.}
    \label{fig:abl_task1}
\end{figure*}

\begin{figure*}[h]
    \centering
\begin{subfigure}{\linewidth}
        \centering
        \includegraphics[width=0.8\linewidth]{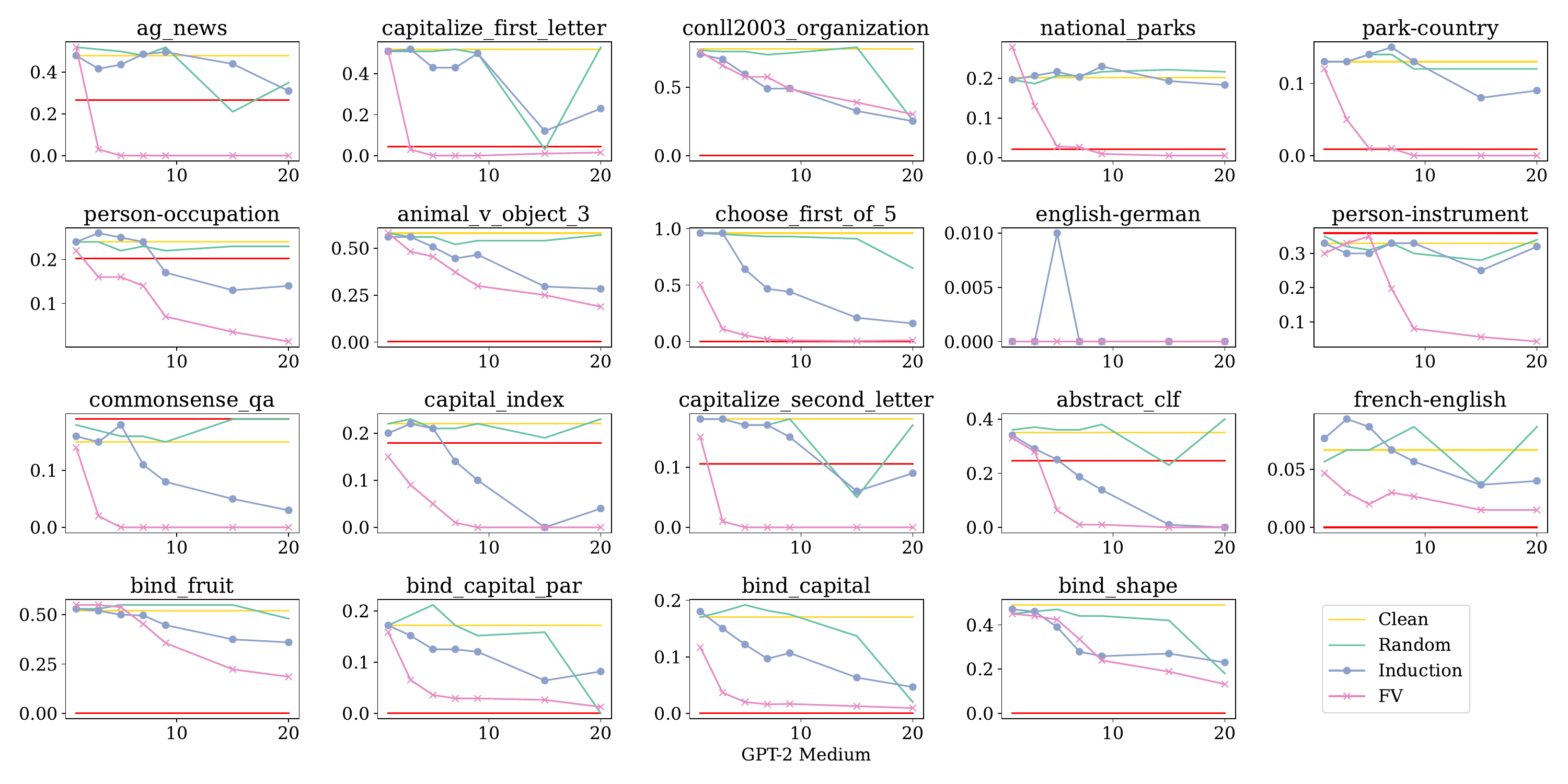} 
    \end{subfigure}
    \vspace{0.5cm}
\begin{subfigure}{\linewidth}
        \centering
        \includegraphics[width=0.8\linewidth]{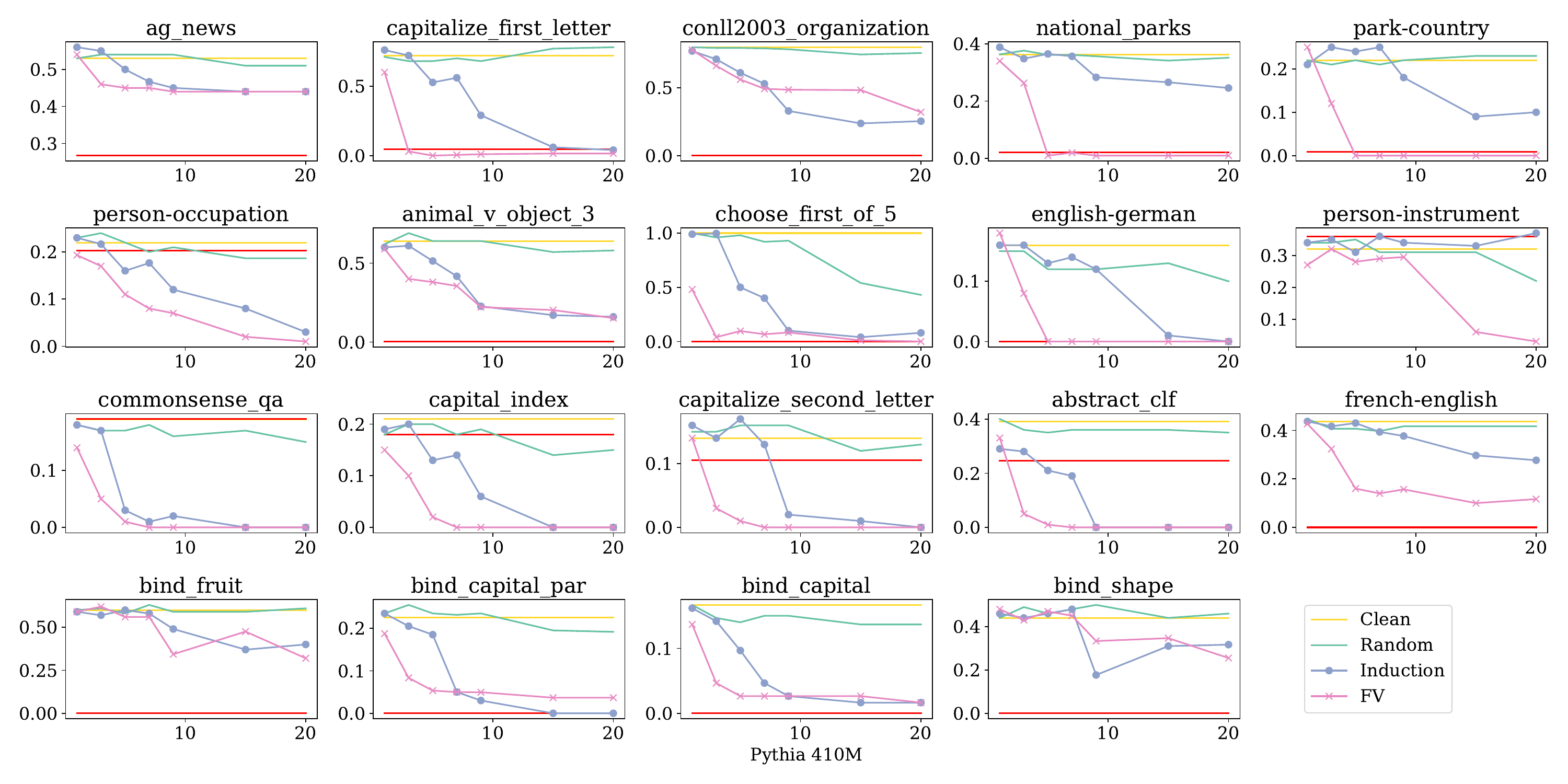} 
    \end{subfigure}
    \vspace{0.5cm}
\begin{subfigure}{\linewidth}
        \centering
        \includegraphics[width=0.8\linewidth]{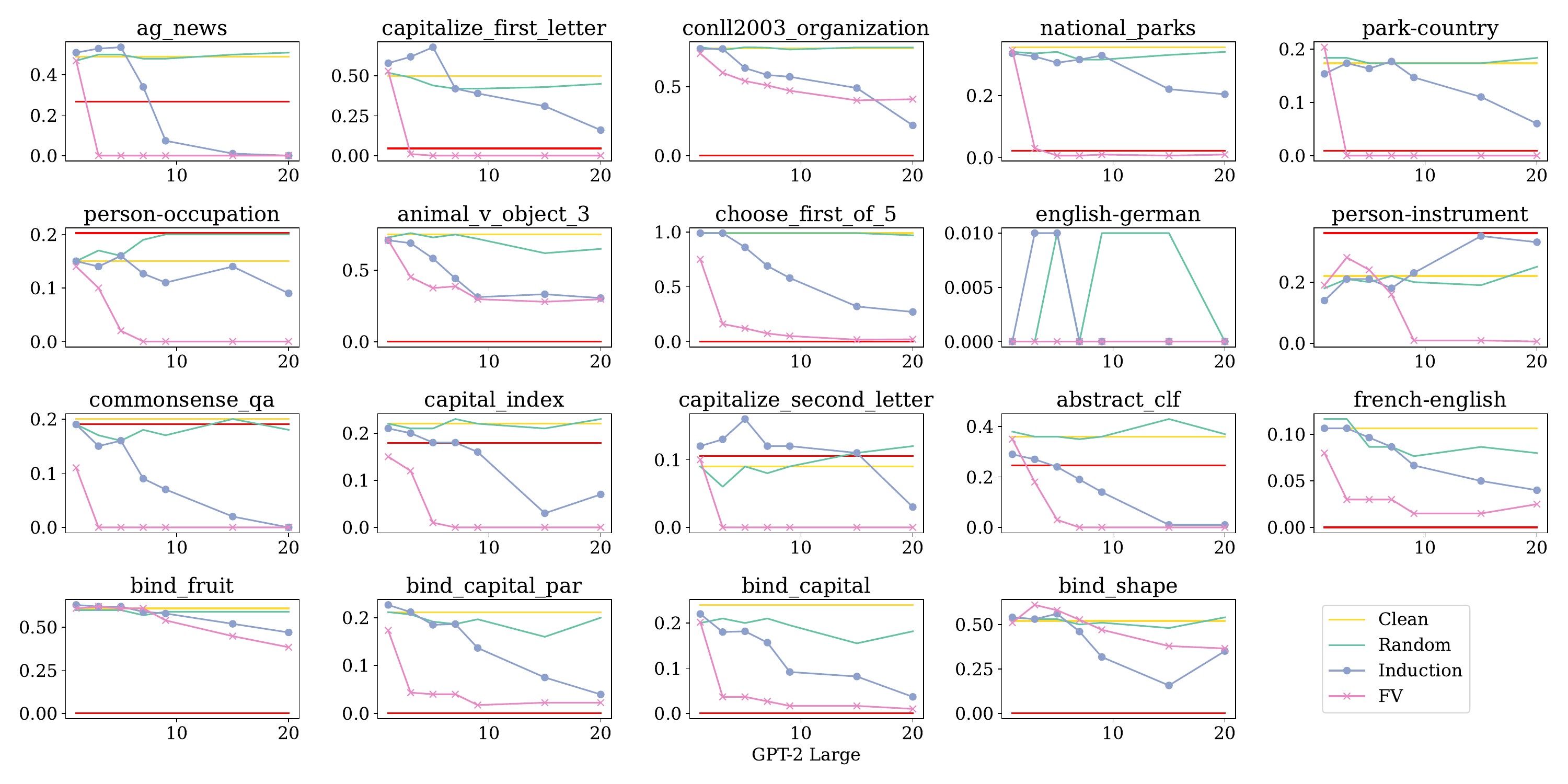} 
    \end{subfigure}
        \caption{ICL accuracy after ablations by task. The red horizontal line represents the random baseline.}
    \label{fig:abl_task2}
\end{figure*}

\begin{figure*}[h]
    \centering
\begin{subfigure}{\linewidth}
        \centering
        \includegraphics[width=0.8\linewidth]{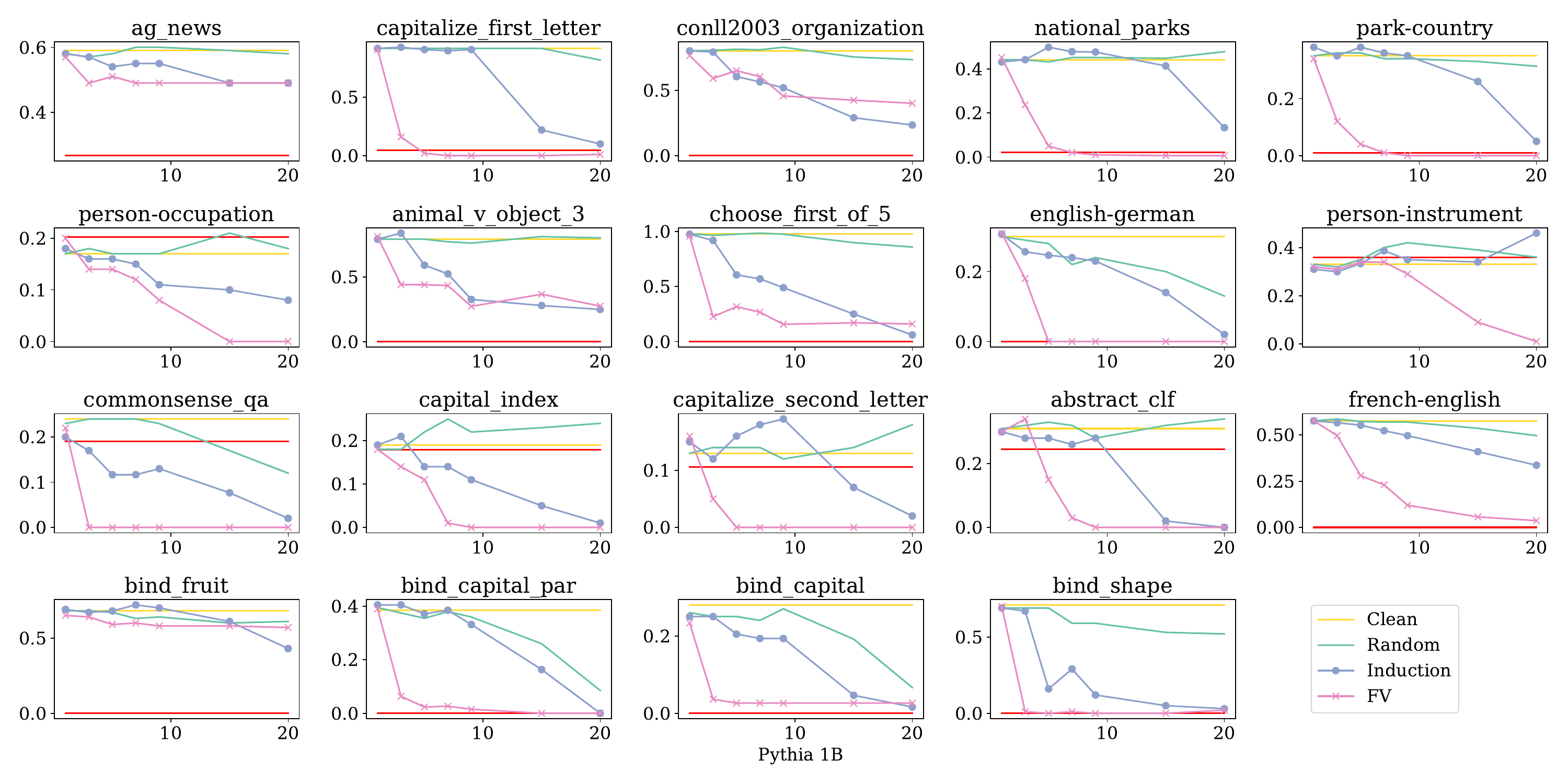} 
    \end{subfigure}
    \vspace{0.5cm}
\begin{subfigure}{\linewidth}
        \centering
        \includegraphics[width=0.8\linewidth]{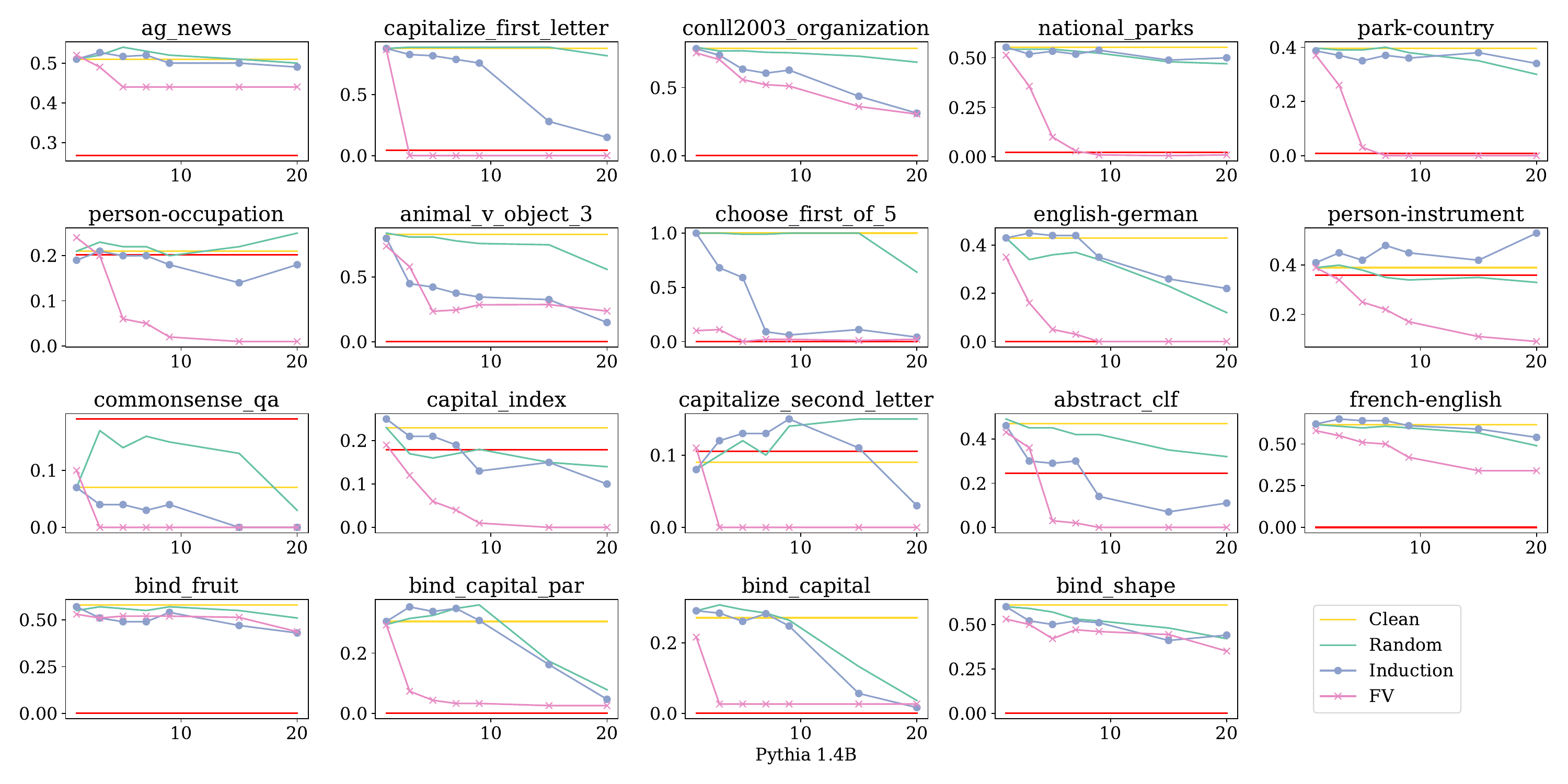} 
    \end{subfigure}
    \vspace{0.5cm}
\begin{subfigure}{\linewidth}
        \centering
        \includegraphics[width=0.8\linewidth]{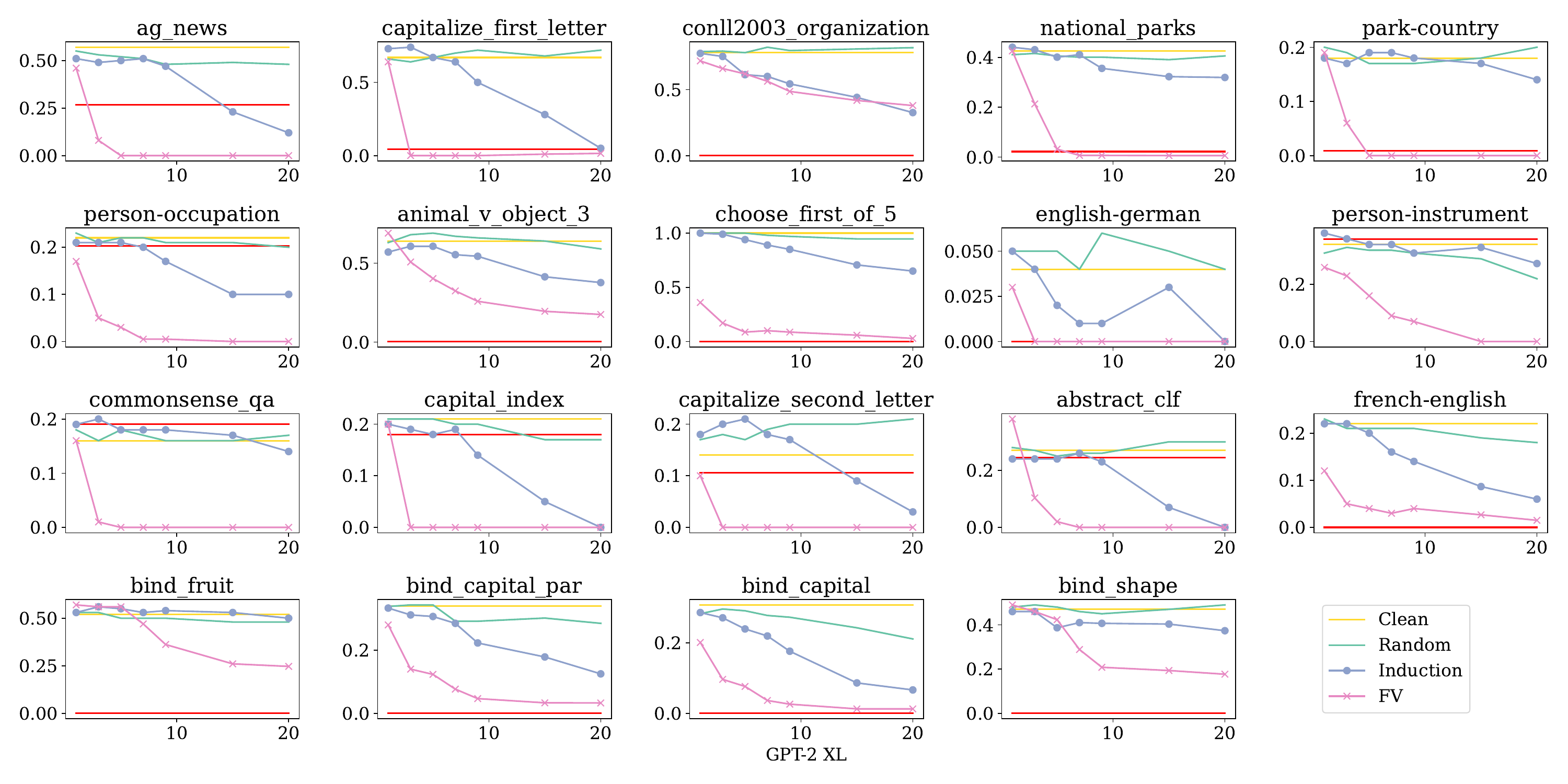} 
    \end{subfigure}
        \caption{ICL accuracy after ablations by task. The red horizontal line represents the random baseline.}
    \label{fig:abl_task3}
\end{figure*}

\begin{figure*}[h]
    \centering\begin{subfigure}{\linewidth}
        \centering
        \includegraphics[width=0.8\linewidth]{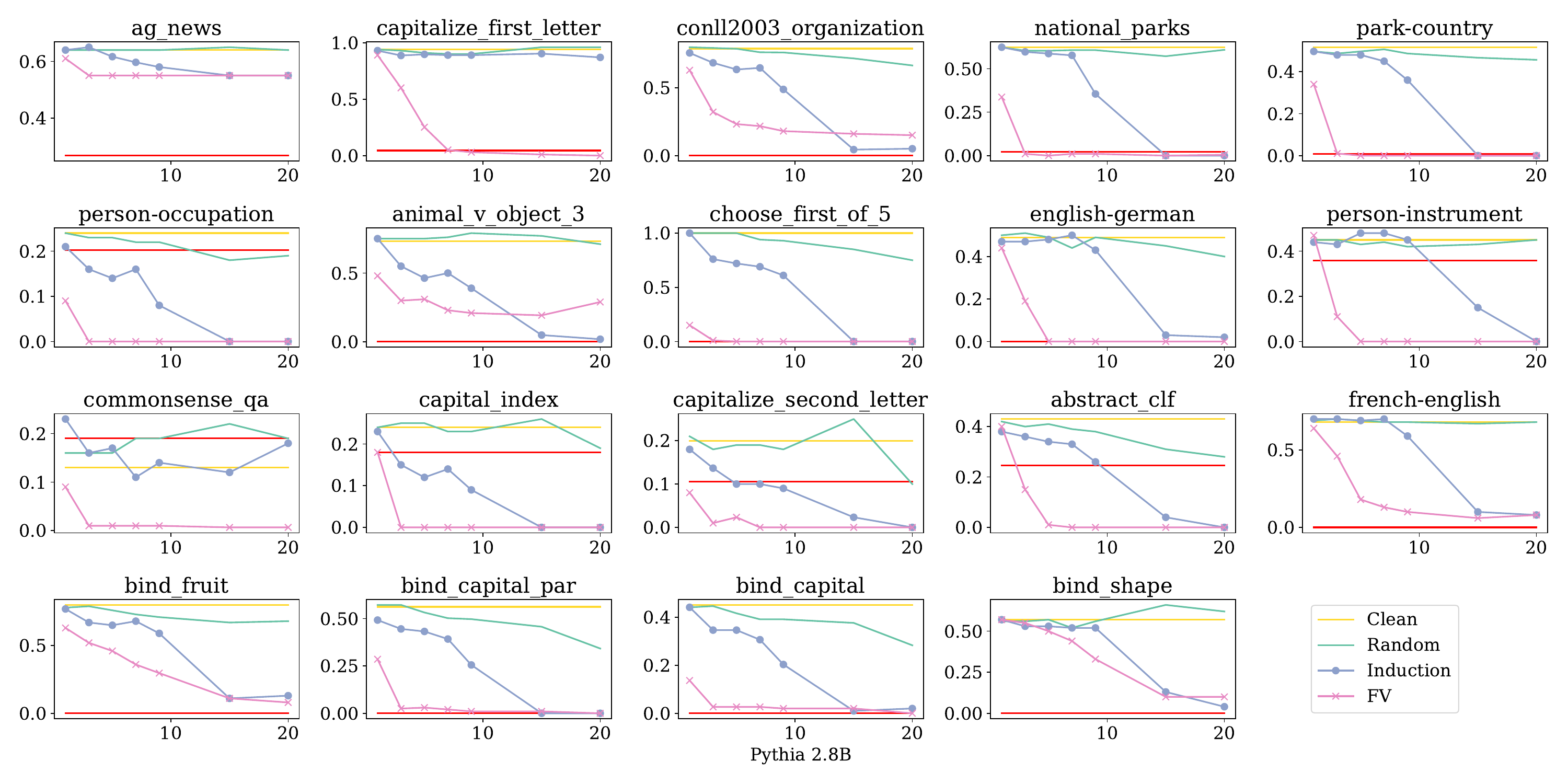} 
    \end{subfigure}
    \vspace{0.5cm}
\begin{subfigure}{\linewidth}
        \centering
        \includegraphics[width=0.8\linewidth]{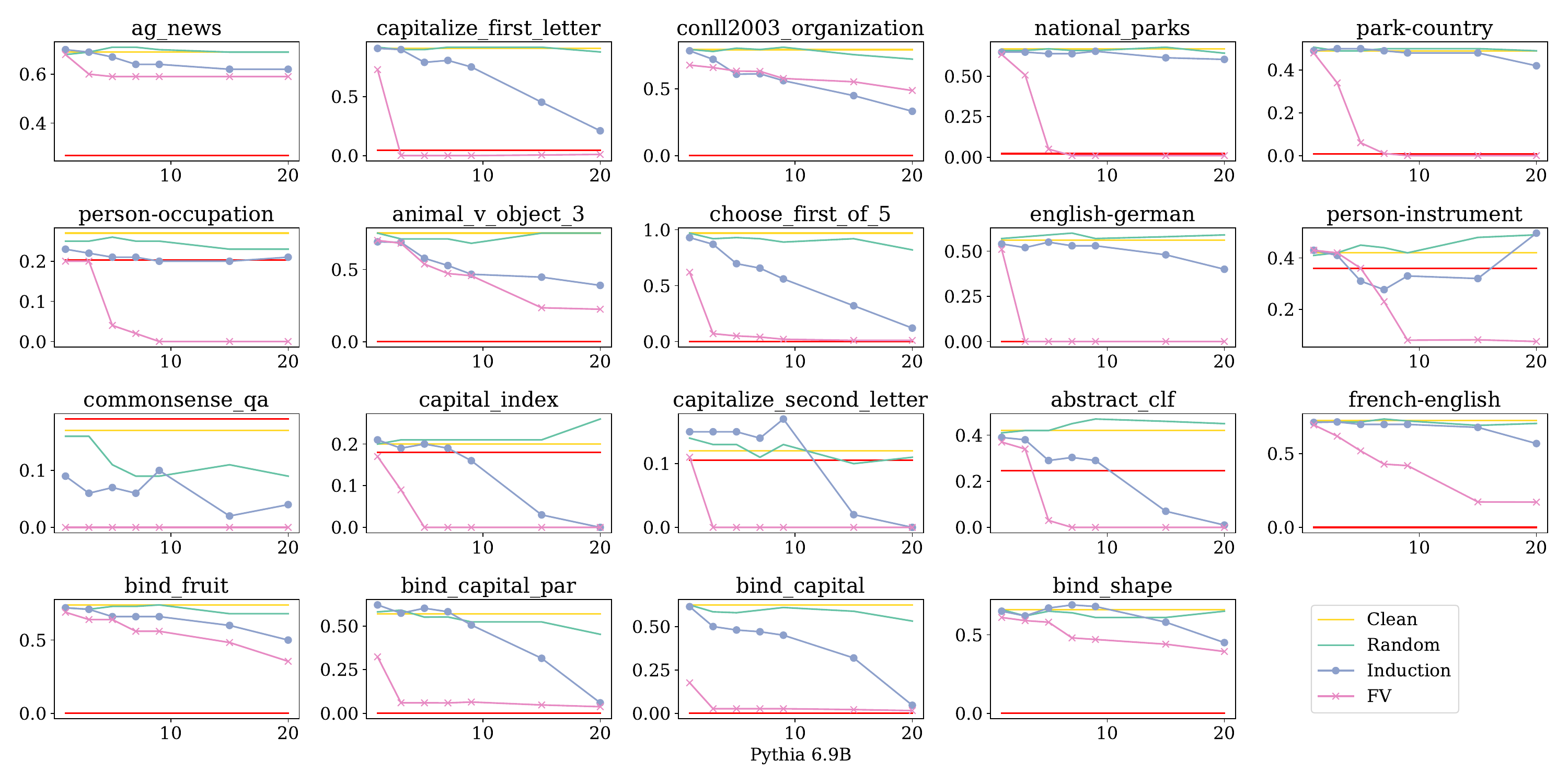} 
    \end{subfigure}
    \vspace{0.5cm}
\begin{subfigure}{\linewidth}
        \centering
        \includegraphics[width=0.8\linewidth]{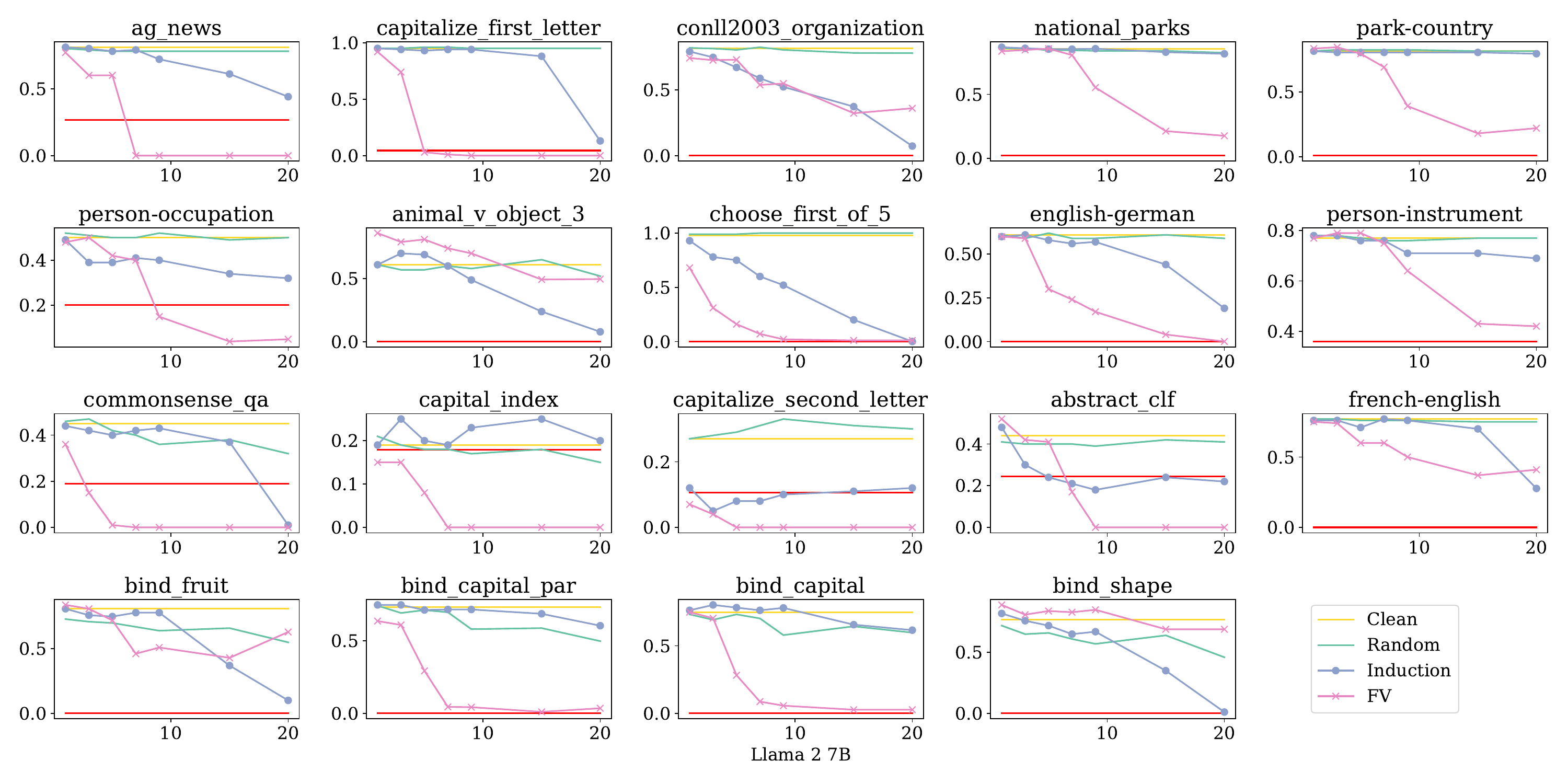} 
    \end{subfigure}
    \caption{ICL accuracy after ablations by task. The red horizontal line represents the random baseline.}
    \label{fig:abl_task4}
\end{figure*}

In Figures \ref{fig:abl_task1}-\ref{fig:abl_task4}, we plot the ICL accuracy after ablating induction heads and FV heads for each task in the evaluation set. We also compute the random baseline for each task, where we randomly sample outputs seen during training and compare these random outputs to the ground truth. The random baselines are shown in red horizontal lines.


\subsection{Head locations}\label{app:layers}
\begin{figure*}
\centering
    \includegraphics[width=\linewidth]{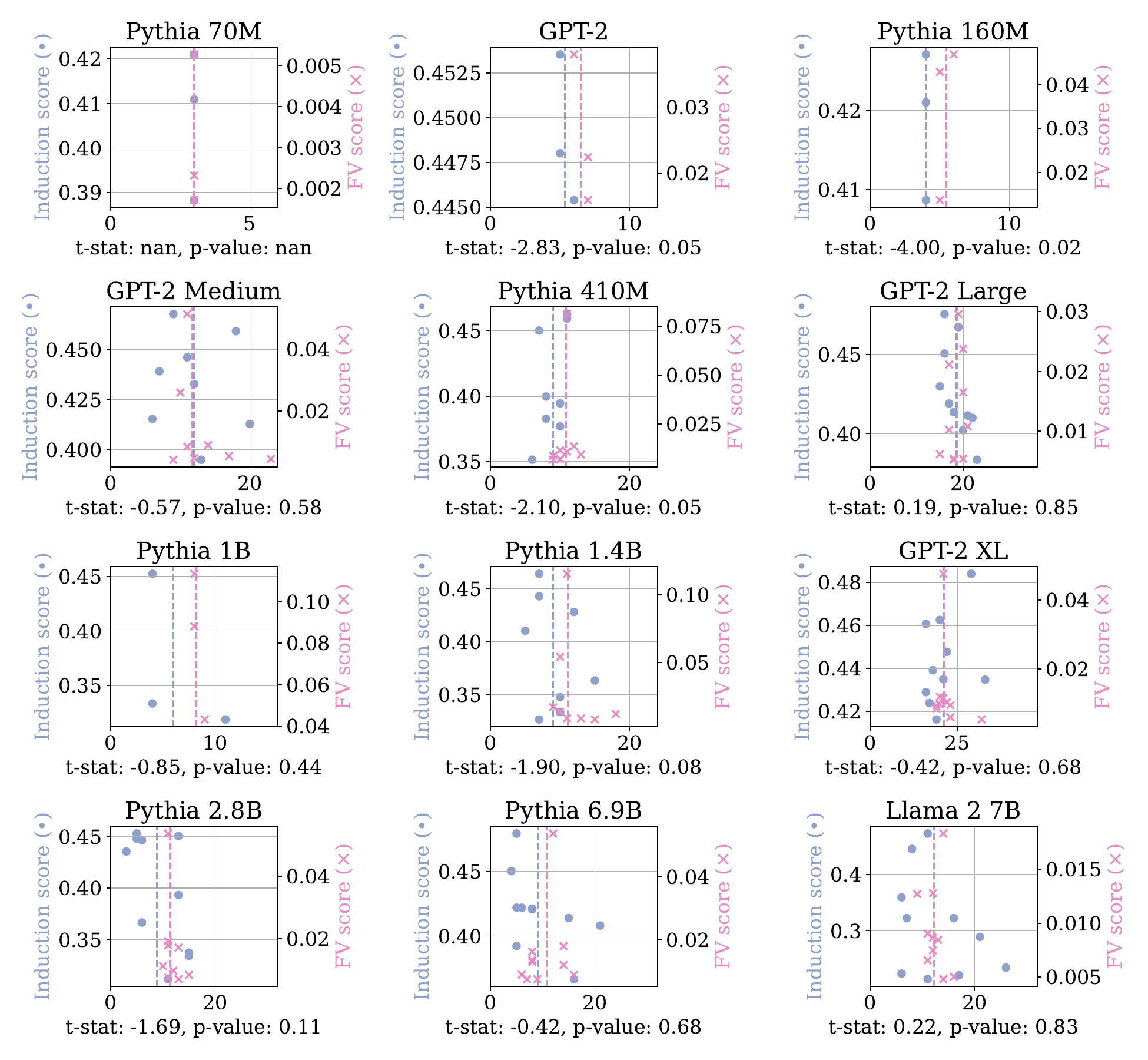}
  \caption{Location of induction heads (blue) and FV heads (pink) in model layers}
  \label{fig:head_layers_all}
\end{figure*}

In Figure \ref{fig:head_layers_all}, we plot the locations of induction heads and FV heads across model layers.

\subsection{Overlap between ablated induction and FV heads}\label{app:overlap}
\begin{figure*}
\centering
    \includegraphics[width=\linewidth]{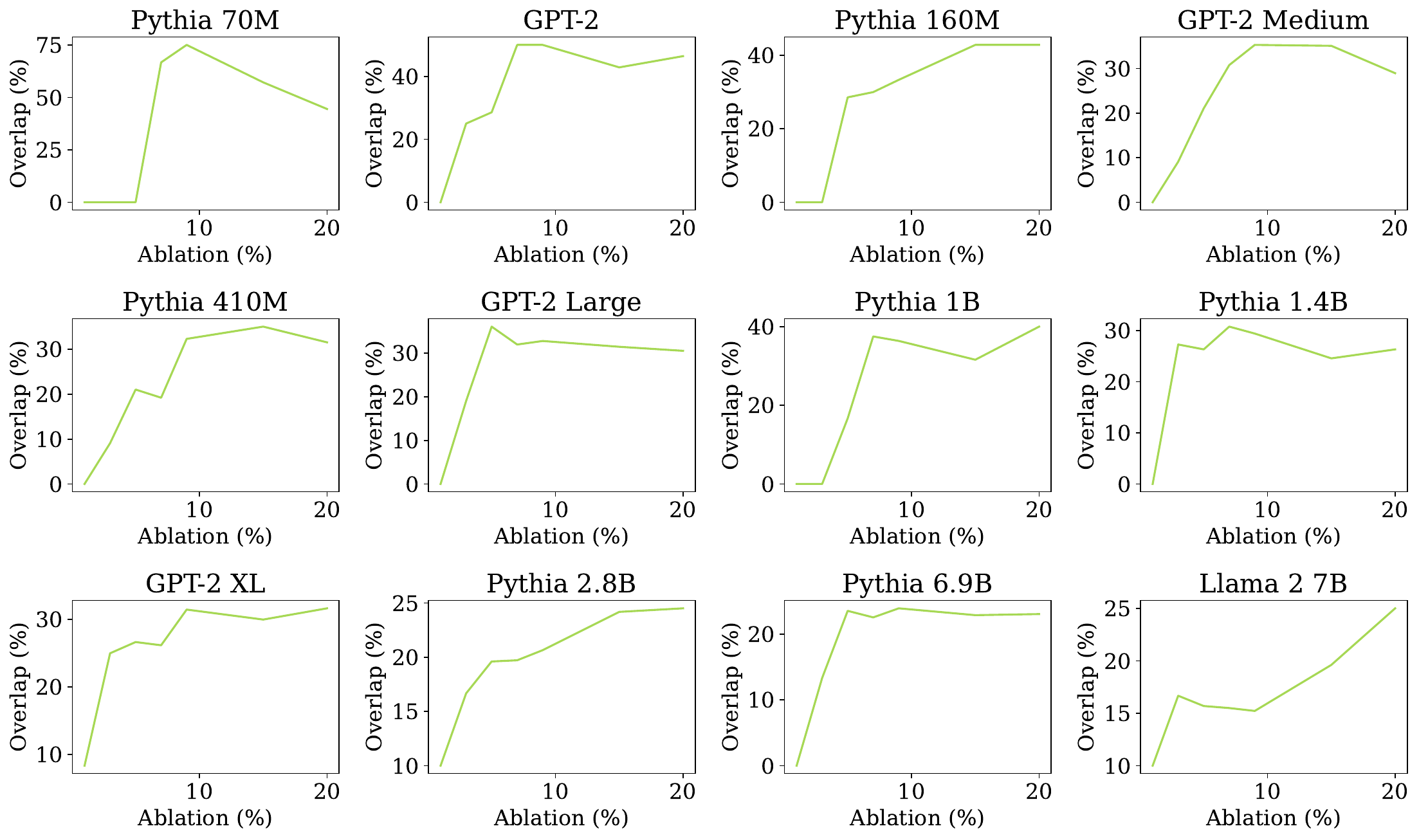}
  \caption{Overlap between set of induction heads and FV heads ablated.}
  \label{fig:abl_overlap}
\end{figure*}

In Figure \ref{fig:abl_overlap}, we plot the percentage of attention heads that overlap between the set of induction heads and FV heads we ablate. We find that as the number of ablated heads increases, the overlap between the two sets of ablated heads also increases. This demonstrates the importance of performing ablations with exclusion to control for overlap.

\subsection{Scores across training}\label{app:train_scores}
\begin{figure*}
\centering
    \includegraphics[width=\linewidth]{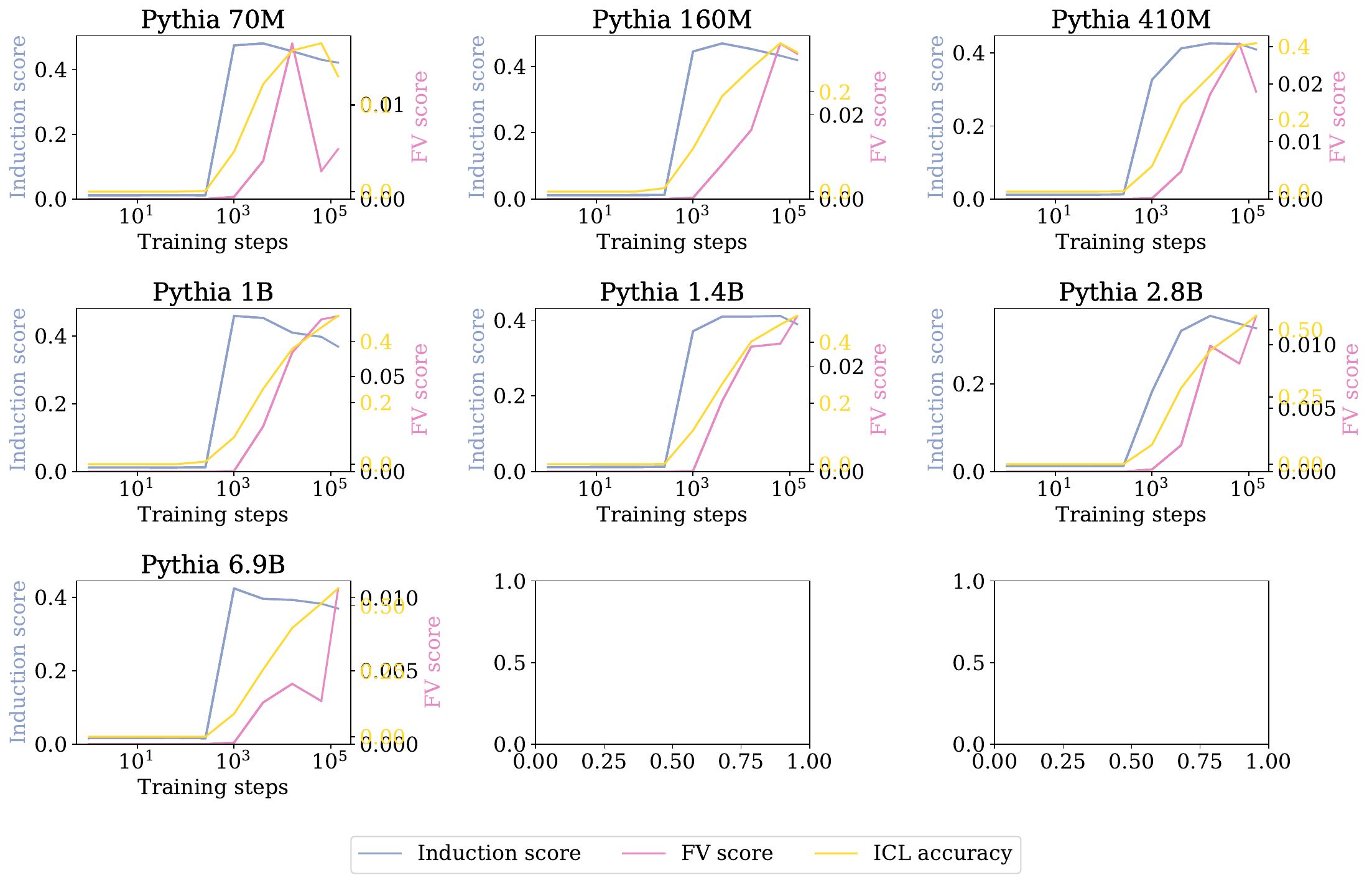}
  \caption{Evolution of induction score and FV score averaged over top 2\% heads across training.}
  \label{fig:mean_ckpt_all}
\end{figure*}

\begin{figure*}
\centering
    \includegraphics[width=0.75\linewidth]{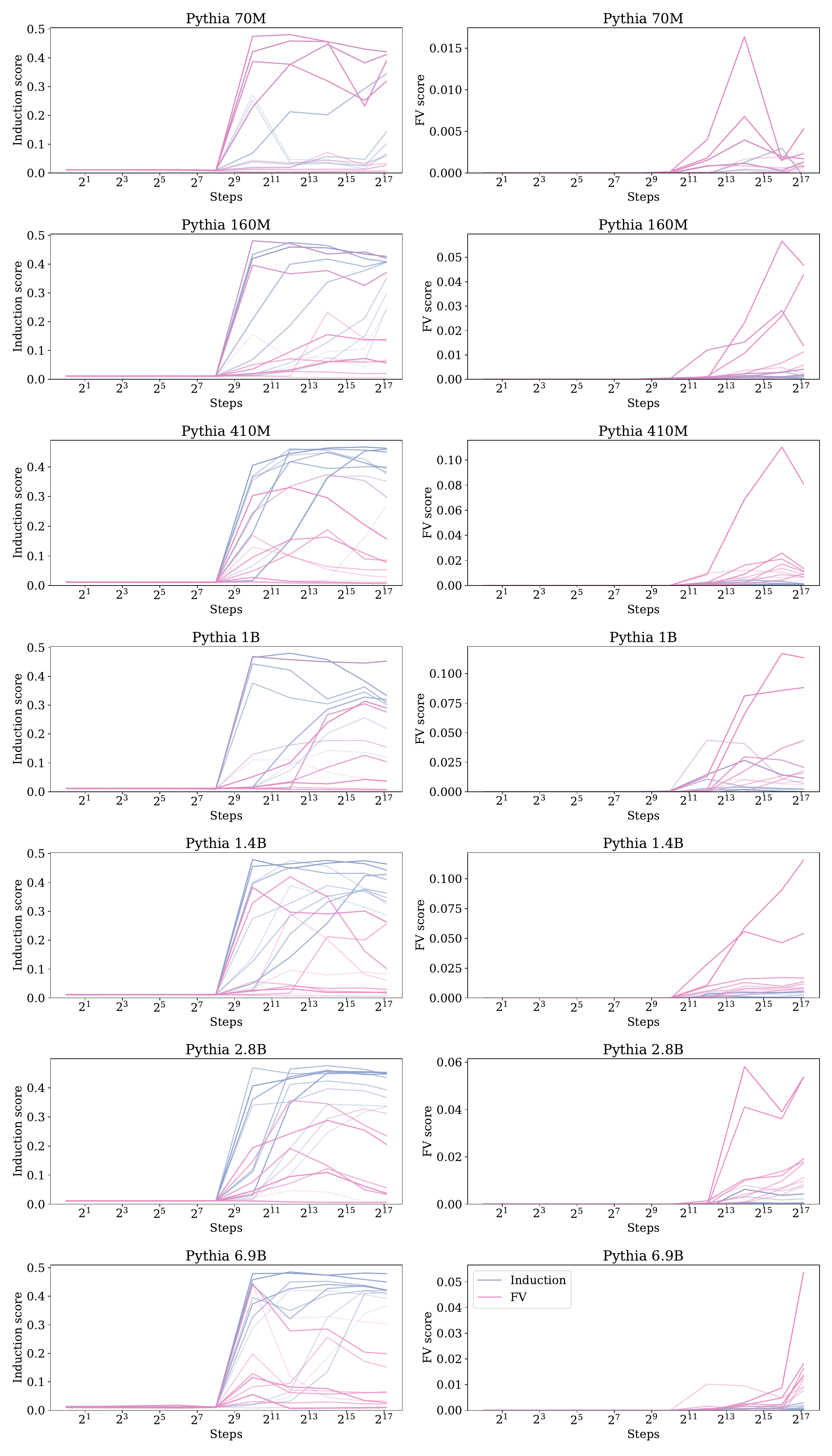}
  \caption{Evolution of induction scores (left) and FV scores (right) of individual induction and FV
heads across training}
  \label{fig:indv_ckpt_all}
\end{figure*}

In Figure \ref{fig:mean_ckpt_all}, we plot the evolution of induction and FV scores averaged over top 2\% heads across model training, along with the few-shot accuracy of the model checkpoints averaged over the evaluation tasks.
In Figure \ref{fig:indv_ckpt_all}, we plot the evolution of induction and FV scores of individual heads across training.

\end{document}